\pdfoutput=1
\documentclass[11pt]{article}

\usepackage[preprint]{acl}

\usepackage{times}
\usepackage{latexsym}
\usepackage{amsmath} 

\usepackage{algorithm}
\usepackage{algpseudocode}

\usepackage{listings}
\usepackage{geometry}
\usepackage[utf8]{inputenc}

\usepackage[T1]{fontenc}
\usepackage[utf8]{inputenc}

\usepackage{microtype}

\usepackage{inconsolata}

\usepackage{graphicx}
\usepackage{booktabs}       % professional-quality tables
\usepackage{amsfonts}       % blackboard math symbols
\usepackage{nicefrac}       % compact symbols for 1/2, etc.
\usepackage{multirow}
\usepackage{float}
\usepackage{hyperref}
\usepackage{url}            

\title{SSFF: Investigating LLM Predictive Capabilities for Startup Success through a Multi-Agent Framework with Enhanced Explainability and Performance}

\author{
  Xisen Wang$^{1}$\thanks{Corresponding author} \\
  \texttt{xisen.wang@keble.ox.ac.uk}
  \And
  Yigit Ihlamur$^{2}$ \\
  \texttt{yigit@vela.partners}
  \And
  Fuat Alican$^{2}$ \\
  \texttt{fuat@vela.partners}
  \AND
  $^{1}$University of Oxford, Oxford, UK \\
  $^{2}$Vela Partners, San Francisco, USA
}

\begin{document}
\maketitle

\begin{abstract}
LLM-based agents have recently demonstrated strong potential in automating complex tasks, yet accurately predicting startup success remains an open challenge with few benchmarks and tailored frameworks. To address these limitations, we propose the Startup Success Forecasting Framework (SSFF), an autonomous system that emulates the reasoning of venture capital analysts through a multi-agent collaboration model. SSFF integrates traditional machine learning models—such as random forests and neural networks—within a retrieval-augmented generation (RAG) framework composed of three interconnected modules: a prediction block, an analysis block, and an external knowledge block. Our evaluations reveal three key findings: 1) by leveraging founder segmentation, we show that startups with L5 founders are 3.79 times more likely to succeed than those with L1 founders; 2) we find that baseline LLMs consistently overpredict startup success and struggle under realistic class imbalances, largely due to an overreliance on founder claims; 3) SSFF significantly enhances prediction accuracy—yielding a 108.3\% relative improvement over GPT-4o-mini and a 30.8\% relative improvement over GPT-4o —thereby mitigating the limitations of standard LLM-based approaches. 

\end{abstract}

\section{Introduction}

Evaluating startups at their inception is a complex task traditionally dependent on the expertise of seasoned venture capital (VC) analysts. The inherent dynamism of startups, coupled with the unpredictable nature of market reception, makes it particularly challenging to identify ventures poised for success. Although recent advances in large language models (LLMs) have opened new avenues for automating complex decision-making processes, these models often suffer from issues such as hallucination effects, overgeneralizations, and 'fuzzy' semantic representations—factors that limit their predictive reliability in real-world scenarios. Moreover, there is currently no widely accepted benchmark for startup evaluation, and little is known about the true predictive capabilities of LLMs in the domain.

BIn this paper, we introduce the Startup Success Forecasting Framework (SSFF), a pioneering multi-agent system that synergizes conventional machine learning methods with the capabilities of LLMs to enhance the evaluation of early-stage startups. SSFF leverages external information retrieval, neural networks, and random forests alongside LLM reasoning to deliver high-quality, data-driven analyses from minimal input data. Our approach explores how LLMs behave in the startup evaluation process and uncovers a pronounced over-prediction bias, where LLMs tend to overestimate success, likely due to overreliance on founder claims. Additionally, we propose and validate a novel founder segmentation method that quantitatively demonstrates the significant impact of founder backgrounds on startup outcomes.

We detail the three key blocks of SSFF—prediction, analysis, and external knowledge—that collectively provide a comprehensive evaluation of startups. The prediction block employs a combination of traditional machine learning techniques and LLM reasoning to generate initial, data-driven predictions from startup and founder information. The analysis block refines these predictions through qualitative assessments and expert-driven segmentation metrics, allowing us to interpret underlying factors more clearly. Meanwhile, the external knowledge block enriches the evaluation by retrieving real-time market and product insights using advanced data extraction and natural language processing techniques. Our framework not only enhances predictive performance and interpretability compared to standalone LLMs but also uncovers novel insights into LLM behavior, such as their propensity to overpredict success in data-constrained settings. The following sections describe our methodology, present both quantitative and qualitative results, and discuss the implications of our findings for future research in VC decision support. Further technical details and in-depth analyses are provided in the Appendix.

\section{Literature Review}

\subsection{Startup Evaluation Pipeline}

Evaluating a startup requires a deep understanding of the startup's market and technology. Startups often have no history, minimal revenue, and low survival rates \citep{Damodaran2009}. The high-risk, high-return nature of venture capital leads to few investors achieving impressive results \citep{Gornall2018,Corea2021}. Traditionally, this process has relied on venture capitalists' intuition, which is inefficient and biased \citep{Astebro2006,Corea2021}. Recent trends favor data-driven approaches, using advanced analytics and machine learning to quantify success factors \citep{Corea2021}.

In recent years, there has been more research around the venture capital and startup ecosystem to improve the performance of startup investing. Xiong (2023) introduced the Founder GPT framework, which uses LLMs to evaluate the "founder-idea" fit, showing that personalized evaluation of a founder's idea alongside her background is crucial to predict startup success \citep{Xiong2023}. In addition, Ozince (2024) showed that LLMs can be effective in segmenting and labeling founders to improve the quality of machine learning methods to predict the success of startups \citep{Ozince2024}. Gavrilenko (2023) and Maarouf (2024) demonstrate how free-form text description of startups can be predictive of future success, while Potanin (2023) provides a high-performance model that can deliver 14 times capital growth \citep{Gavrilenko2023, Potanin2023, Maarouf2024}. 

\subsection{LLM Agent}

The advent of LLMs such as GPT (Generative Pre-Trained Transformer) and BERT (Bidirectional Encoder Representations from Transformers) has revolutionized natural language understanding, showing transformative potential across various research fields \citep{Brown2020, Touvron2023, Zhu2024}. As foundational platforms, LLMs have expanded the concept of AI agents, defined as systems that perceive and respond to environmental data, producing meaningful actions \citep{Durante2024}. Role-Play support systems, for example, enable AI agents to become increasingly human-like in diverse scenarios \citep{Shanahan2023}. These agents can autonomously analyze data and support decision-making, identifying complex patterns beyond human capabilities. Despite these advancements, a widely-adopted AI agent specifically for startup analysis has yet to emerge.

\subsection{Prompting Techniques}
The effectiveness of LLMs in various applications depends heavily on prompt engineering, which involves crafting queries to guide the model for specific outputs. Well-designed prompts enhance LLM performance, making this skill crucial for startup evaluation. Techniques like Chain-of-Thought, Tree-of-Thought, Few-shot Learning, and Retrieval Augmented Generation improve AI's accuracy and data retrieval \citep{Wei2022, Yao2023, Srinivasan2022, Zhu2024}. This paper employs these techniques in a divide-and-conquer framework to improve AI's utility and reliability in problem-solving, strategic planning, information retrieval, and decision-making.

\section{Founder Segmentation}

\subsection{Data Preparation}
A comprehensive dataset of over 2000 entries was curated from verified LinkedIn profiles, capturing only the historical data available up to each startup’s founding or VC involvement to ensure an authentic reflection of founders’ early backgrounds. Detailed records include educational credentials, work experiences, and key roles, and are further enriched with startup composition data that tracks changes across life stages—providing essential context on team dynamics and early-stage structures. Startups are classified as “successful” or “unsuccessful” based on market valuations (with those exceeding USD 500 million deemed successful) and augmented with supplementary insights from Crunchbase, a paid data source offering deeper information on founders’ professional and educational histories (subject to licensing restrictions). For our segmentation analysis, 1000 entries from each label were randomly sampled, forming a robust foundation for evaluating founder segmentation.

To embark on the analysis, we initially curated a dataset comprising founders' LinkedIn profiles, associated with startups classified as either successful or unsuccessful. This classification was based on the companies' market valuations, with successful ones having valuations over USD 500 million. The dataset was enriched with detailed profiles, including education and work backgrounds, extracted in JSON format from LinkedIn URLs. This preparation phase was crucial for ensuring a robust foundation for our segmentation analysis. Both labels contained more than 2000 entries of founders and their respective companies. After processing, 1000 entries of founders, sampled randomly from each label, were used for segmentation. 

\subsection{Segmentation Process}
Founders were segmented into five levels—from Level 1 (least qualified) to Level 5 (most qualified)—using GPT-4 to evaluate their LinkedIn profiles. The segmentation criteria, informed by years of VC expertise, focus on key indicators such as leadership roles, business achievements, and educational background. Tailored prompts were iteratively refined to ensure accurate and nuanced categorization. Additional details—including the prompt, criteria, and sample segmentation results—are provided in the Appendix.

\subsection{Segmentation Results}
The segmentation results demonstrate a strong correlation between founder levels and startup success rates. Founders at Level 5 (L5), characterized by their proven track record in building significant businesses or holding executive roles at leading technology companies, were substantially more likely to lead their startups to success. Specifically, Level 2 (L2) founders were 1.12 times more likely to succeed than Level 1 (L1) founders, while Level 5 founders were an impressive 3.79 times more likely to succeed than their L1 counterparts.

\begin{table}[ht]
\centering
\caption{Success and failure rates by founder level, showcasing the predictive power of founder segmentation on startup success.}
\begin{tabular}{@{}lcccc@{}}
\toprule
Level & Success & Failure & Success Rate \\
\midrule
L1 & 24 & 75 & 24.24\% \\
L2 & 83 & 223 & 27.12\% \\
L3 & 287 & 445 & 39.21\% \\
L4 & 514 & 249 & 67.37\% \\
L5 & 93 & 8 & 92.08\% & \\
\bottomrule
\end{tabular}
\label{tab:founder_segmentation_results}
\end{table}

These findings highlight the significant impact of a founder’s background on startup success and suggest that incorporating this nuanced segmentation into evaluation frameworks can enhance predictive accuracy. Moreover, exploring even more granular segmentation may further refine these predictions

\section{Startup Success Forecasting Framework} 

\subsection{Analyst Block}

\paragraph{Divide-And-Conquer Approach}

The Analyst Block of the SSFF is a critical component in evaluating startup ecosystems. It operates using a multi-agent framework, where startup data is processed and filtered through a divide-and-conquer strategy. Each agent receives specific, relevant portions of the information and works independently to analyze different aspects of the startup.

The VC-Scout Agent initiates this process by categorizing the startup data into 18 dimensions, covering various aspects such as general descriptions, growth rates, and regulatory approvals. Each specialized agent is assigned a dimension to investigate, acting as an expert in its respective domain.

\begin{lstlisting}
{
  "startup_analysis_responses": {
    "name": "Startup ABC",
    "description": "Revolutionizing the fintech industry with a blockchain-based payment solution.",
    "market_size": "Large",
    "growth_rate": "20% annually",
    ...
    "regulatory_approvals": "Compliant with all applicable fintech regulations",
    "patents": "2 patents on blockchain transaction algorithms"
  }
}
\end{lstlisting}

\paragraph{Multi-Agent Framework}

The multi-agent framework simulates a team of venture capital analysts, with each agent focusing on different areas of startup evaluation. After the VC Scout processes the initial data, each agent works independently to assess its part of the analysis. An "Integration Agent" then synthesizes these insights into a final, cohesive evaluation.

The main agents involved in the evaluation process are as follows:

\begin{enumerate}
    \item \textbf{Market Agent:} Evaluates the startup's market alignment, growth potential, and strategic positioning.
    \item \textbf{Product Agent:} Assesses the innovation, scalability, and user engagement to determine product-market fit.
    \item \textbf{Founder Agent:} Examines the founding team's expertise, background, and long-term vision.
\end{enumerate}

This multi-agent approach ensures that the startup is comprehensively assessed across all critical areas, with each agent providing a specialized, in-depth analysis, culminating in a well-rounded evaluation.

\begin{figure}
    \centering
    \includegraphics[width=1\linewidth]{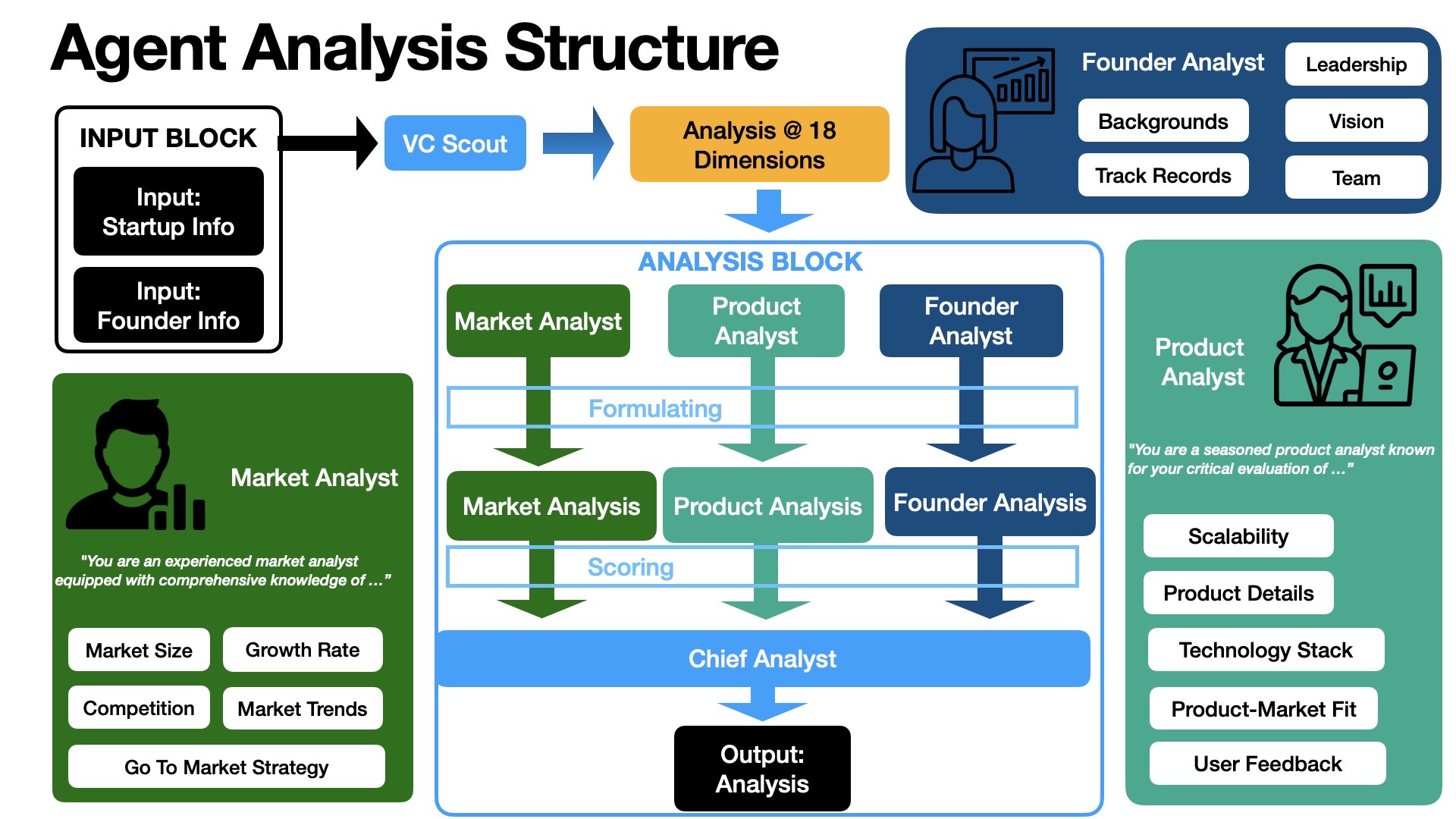}
    \caption{Overview of the Mult-Agent Analyst Block}
    \label{fig:enter-label}
\end{figure}

\subsection{Prediction Block}

A prediction block is designed to learn from past data and predict the likelihood of success as numerical metrics to support the reliability of an automated startup evaluation pipeline. The prediction block is separated into two parts: (1) LLM-Based Random Forest Model and (2) Founder-Idea Fit Network. The preparation of data, relevant parameters, and definitions of labels are explained in the Appendix under the Founder Segmentation section for brevity. 

\subsubsection{LLM-Based Random Forest}

\paragraph{Model Design}

The conventional Random Forest algorithm, celebrated for its effectiveness and explainability, often faces challenges with categorical variables due to its inherent design constraints. To overcome these limitations, we introduce an LLM-based Random Forest model. This novel approach utilizes LLMs, particularly GPT-4o, for the extraction of features, imbuing the model with the flexibility to handle a broad spectrum of categorical variables.

In our framework, startup and founder information is processed through an LLM to categorize data across 14 dimensions, including industry growth, market size, development pace, and product-market fit, among others. This method allows for a nuanced understanding of startup dynamics, which is critical for accurate prediction. The sample data analyzed by the LLM is structured as follows:

\begin{verbatim}
{
  "startup_analysis_responses": {
    "industry_growth": "Yes",
    "market_size": "Large",
    "development_pace": "Faster",
    ...
    "timing": "Just Right"
  }
}
\end{verbatim}

The mathematical formulation of the algorithm is summarized as a pseudo-algorithm in the Appendix. 

This approach to leveraging LLMs for feature extraction and encoding provides a flexible and robust framework for startup success prediction, making full use of categorical variables without the constraints of traditional models.

\begin{algorithm}
\caption{LLM-enhanced Random Forest Model}\label{alg:LLM-RF}
\begin{algorithmic}[1]
    \small
    \Require Dataset \( \mathcal{D} = \{(x_i, y_i)\}_{i=1}^N \)
    \Ensure Trained Random Forest model \( \mathcal{F}_{\text{RF}} \) 
    \State Define features \( \{f_1, f_2, \ldots, f_{14}\} \) with corresponding outcomes
    \State Use LLM to categorize data \( A \) into \( \{c_1, c_2, \ldots, c_N\} \).
    \State Encode categorical features into numerical values.
    \State Split dataset \( \mathcal{D} \) into training set \( \mathcal{D}_{\text{train}} \) and testing set \( \mathcal{D}_{\text{test}} \).
    \[
    \mathcal{D}_{\text{train}} = \{(x_i, y_i)\}_{i=1}^{N_{\text{train}}}, \quad \mathcal{D}_{\text{test}} = \{(x_i, y_i)\}_{i=N_{\text{train}}+1}^{N}
    \]
    where \( N_{\text{train}} + N_{\text{test}} = N \).
    \State Train Random Forest model \( \mathcal{F}_{\text{RF}} \) on \( \mathbf{X}_{\text{train}} \) and \( \mathbf{Y}_{\text{train}} \).
    \[
    \Theta = \arg\min_{\Theta} \mathcal{L}(\mathcal{F}_{\text{RF}}(\mathbf{X}_{\text{train}}; \Theta), \mathbf{Y}_{\text{train}})
    \]
    \State Predict \( \mathbf{Y}_{\text{test}} \) using trained model \( \mathcal{F}_{\text{RF}} \).
    \[
    \hat{\mathbf{Y}}_{\text{test}} = \mathcal{F}_{\text{RF}}(\mathbf{X}_{\text{test}}; \Theta)
    \]
    \State Calculate accuracy of predictions.
    \[
    \text{Accuracy} = \frac{1}{N_{\text{test}}} \sum_{i=1}^{N_{\text{test}}} \mathbb{1}(\hat{y}_i = y_i)
    \]
    where \( \mathbb{1}(\cdot) \) is the indicator function.
\end{algorithmic}
\end{algorithm}

\textbf{Note:} The full prompt and methodology, including question design and LLM interaction, are attached to the appendix for further investigation.

\paragraph{LLM-Based Categorical Data Extraction}
A cornerstone of our LLM-based Random Forest model is the extraction of categorical data from startup and founder information. This process is guided by a Chain of Thought prompting technique, where the LLM is presented with a series of questions designed to elicit specific insights into various aspects of a startup's potential for success. These questions cover a wide range of topics for comprehensiveness. 

Some illustrative questions used in this process are as follows:

\begin{enumerate}
    \item "Is the startup operating in an industry experiencing growth? [Yes/No/N/A]"
    \item "Is the target market size for the startup's product/service considered large? [Small/Medium/Large/N/A]"
    \item "Does the startup demonstrate a fast pace of development compared to competitors? [Slower/Same/Faster/N/A]"
\end{enumerate}

The corresponding encoding is presented in the table below.

\begin{table}[htbp]
\centering
\caption{Adjusted Category Mappings with "Mismatch" Included}
\label{tab:category_mappings}
\begin{tabular}{p{3cm} p{4cm}}
\hline
\textbf{Category} & \textbf{Mappings} \\
\hline
Industry Growth & No, N/A, Yes, Mismatch \\
Market Size & Small, Medium, Large, N/A, Mismatch \\
Development Pace & Slower, Same, Faster, N/A, Mismatch \\
...... \\
\hline
\end{tabular}
\end{table}

This structured approach to querying provides a rich dataset from which we can extract categorical variables with high relevance to startup success. The responses to these questions are then encoded into numerical values, forming the basis for training our LLM-based Random Forest model. 

\paragraph{Model Performance Evaluation}

Our LLM-based Random Forest model was trained on a random \& balanced subset of 1,400 instances—comprising equal numbers of successful and unsuccessful startups—drawn from our curated dataset. For feature extraction and model guidance, we employed GPT-4o. Table~\ref{tab:classification_report} summarizes the classification performance of our model on the test set.

\begin{table}[ht]
\centering
\caption{Classification report for the model.}
\label{tab:classification_report}
\small
\begin{tabular}{lcccc}
\hline
Class & Precision & Recall & F1-Score & Support \\ 
\hline
0 & 0.79 & 0.72 & 0.75 & 137 \\
1 & 0.75 & 0.81 & 0.78 & 142 \\
\hline
Accuracy & & & 0.77 & 279 \\
Macro avg & 0.77 & 0.77 & 0.77 & 279 \\
Weighted avg & 0.77 & 0.77 & 0.77 & 279 \\
\hline
\end{tabular}
\end{table}

Notably, training with only 200 data points using GPT-3.5 yielded an average accuracy of 68\%, whereas scaling up to 1,400 data points with GPT-4o resulted in a performance improvement of approximately 10 percentage points. Moreover, when evaluated under a realistic, skewed distribution with a 1:4 ratio of successful to unsuccessful startups, the model achieved an accuracy of 80.00\% and an F1 score of 54.55\%. These results highlight not only the robustness of our approach but also the critical impact of training data size and balance on predictive accuracy, underscoring the promise of integrating advanced AI techniques with traditional machine learning models for startup evaluation.

\subsubsection{Founder-Idea Fit Network }

SSFF incorporates a novel network to assess founder-idea fit, which is a crucial aspect of startup success prediction. The Founder-Idea Fit Model is designed to quantitatively assess the alignment between founders' expertise and characteristics and their startup's core idea and market positioning. 

\paragraph{Measuring Founder-Idea Fit}
Our analysis highlights a strong correlation between founder segmentation levels and startup outcomes. For instance, Level 5 founders are over three times as likely to succeed as Level 1 founders. However, exceptions exist, with some Level 5 founders failing and some Level 1 founders succeeding. To capture these nuances, we introduce the Founder-Idea Fit Score (FIFS), a metric that quantifies the compatibility between a founder’s experience and the viability of their startup idea.

We define the preliminary FIFS as:

$FIFS(F, O) = (6 - F) \times O - F \times (1 - O)$

where F represents the founder’s level (ranging from 1 to 5) and O denotes the outcome (1 for success, 0 for failure). To facilitate comparison across startups, the FIFS is normalized to a range of [-1, 1] using the formula:

$\text{Normalized FIFS} = \frac{FIFS}{5}$

On this scale, a score of 1 indicates the optimal founder-idea fit—achieved when a Level 1 founder succeeds—while a score of -1 reflects the poorest fit, as observed when a Level 5 founder fails. This normalized metric enables straightforward comparisons, highlighting startups with the strongest or weakest alignment between founder capabilities and their business concepts.

\paragraph{Preprocessing: Embedding \& Cosine Similarity}

The first step in the Founder-Idea Fit Model is to generate dense vector representations for both startup descriptions and founder backgrounds. We employ OpenAI’s \textit{text-embedding-3-large} model to transform textual data into 100-dimensional embeddings, capturing the semantic essence of each description. These embeddings encapsulate details such as a startup’s mission, technology, and market, as well as a founder’s education and employment history. We then compute the cosine similarity between each founder’s embedding and the corresponding startup’s embedding, using this metric as a proxy for the semantic alignment—or “fit”—between the founder’s background and the startup’s concept.

\paragraph{Statistical Analysis and Further Model Considerations}

Our statistical analysis revealed only a modest linear association between cosine similarity and the Founder-Idea Fit Score (FIFS), with a Pearson correlation coefficient of 0.173 and an R-squared value of 0.030 from an Ordinary Least Squares (OLS) regression. These findings indicate that cosine similarity accounts for merely 3\% of the variability in FIFS. Although statistically significant, the low predictive power, positive autocorrelation, and deviations from normality suggest that a linear model is inadequate. As a result, we developed a neural network to better capture the nonlinear relationships among the features, thereby improving predictive accuracy.

\paragraph{Model Architecture and Performance Analysis}

Our neural network model uses the embeddings and their cosine similarity as input features to predict the Founder-Idea Fit Score. The model features a straightforward architecture with an input dense layer of 128 neurons followed by a second dense layer of 64 neurons, both activated by ReLU and regularized with dropout rates of 20\% and 30\%, respectively. Trained on a dataset of founder-startup pairs, the network achieved a dramatic reduction in mean squared error—from 0.7182 during initial epochs to 0.0407 at convergence—and a validation loss of 0.0386. These results underscore the robustness and scalability for early-stage founder-idea fit evaluation.

\subsection{External Knowledge Block}

The External Knowledge Block is essential to SSFF, as it enriches our analysis with real-time market and product insights. This module leverages a Retrieval-Augmented Generation (RAG) framework that combines web-scraping APIs (such as the SERP API) with advanced natural language processing techniques to systematically gather, filter, and synthesize relevant market data. In practice, the process involves generating targeted keywords, retrieving diverse content—such as news articles, blog posts, and market reports—and distilling this information into structured insights on market size, growth, trends, and consumer sentiment.

For further technical details—such as the specific keywords generated, the hierarchical levels of information retrieved, and a comparative analysis of retrieval performance (e.g., varying the parameter from N=3 to N=10)—please refer to the Appendix.

\subsection{SSFF Framework Design}

\begin{figure}
    \centering
    \includegraphics[width=1\linewidth]{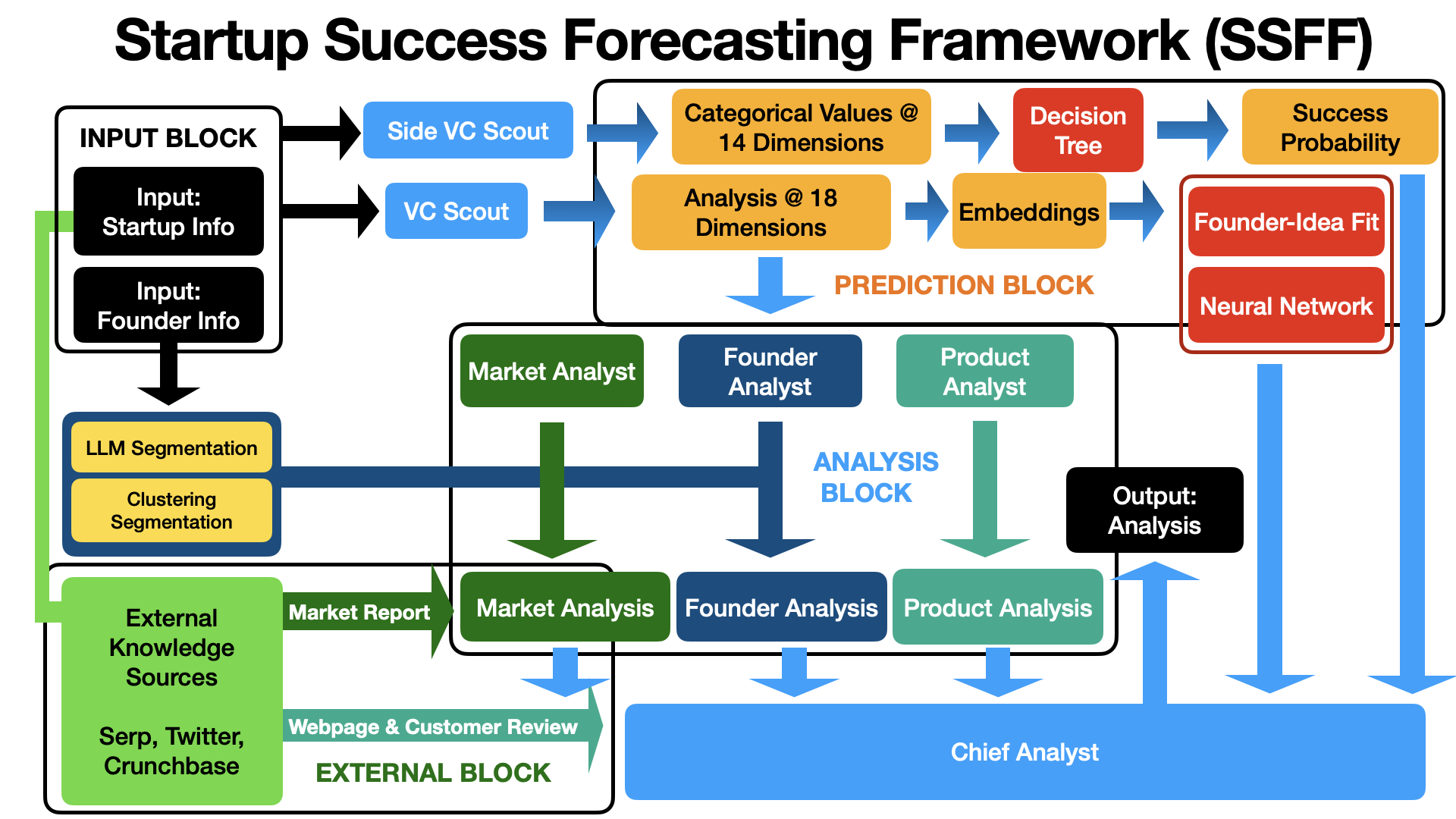}
    \caption{A schematic view of the Startup Success Forecasting Framework}
    \label{fig:enter-label}
\end{figure}

The SSFF framework seamlessly integrates three core blocks to deliver a holistic, data-driven analysis of startup success. It begins with a VC scout agent that scrutinizes founder and startup data across 18 dimensions, establishing a robust analytical foundation. In parallel, specialized analysts decompose the problem further—one branch employing a decision tree model with 14 tailored questions (side VC) to generate initial predictions, and another leveraging the Founder-Idea Fit model to segment and evaluate founders, ultimately computing a Fit Score. Complementing these internal analyses, the External Knowledge Block gathers real-time market intelligence and relevant news through advanced web-scraping and NLP techniques. 

All these insights are then synthesized by a chief analyst, who produces a comprehensive report featuring detailed quantitative scores, predictive outcomes, and a clear rationale for recommendations such as “Invest” or “Hold.” This interconnected design not only improves predictive accuracy but also ensures transparency and interpretability beyond what standalone LLMs can offer.

\section{Evaluations \& Results}

\subsection{Setup}

In our experiment, we selected a stratified sample of 50 startups from our curated database to reflect real-world conditions—40 unsuccessful startups and 10 successful ones. Each startup was represented by minimal input, consisting solely of the relevant founder information and a brief startup description (under 100 words), thereby simulating the limited data typically available in large-scale VC sourcing. Baseline performance was compared using \textbf{GPT-4o} and \textbf{GPT-4o-mini}, focusing our study on assessing how LLMs predict startup success under realistic, data-constrained conditions. 

Our evaluation framework assessed several key parameters, including \textbf{Founder Idea Fit}, \textbf{Categorical Prediction} (as implemented in our LLM-based RF model), \textbf{Quantitative Decision} (derived solely from quantitative metrics such as Founder Idea Fit and associated scores), and \textbf{Decision Confidence} (the confidence level generated by the LLM). In addition, comprehensive analyses were provided via \textbf{Product Report}, \textbf{Market Report}, and \textbf{Founders Report}, which were subsequently synthesized into a \textbf{Final Report}. The framework further incorporates detailed \textbf{Founder Segmentation} (with accompanying reasoning) and culminates in a combination of \textbf{Final Decision}, \textbf{Overall Score}, \textbf{Level of Confidence(Quantitative Probability)} generated by the integration agent after aggregating all insights.

\subsection{Quantitative Evaluations}

\begin{figure}
    \centering
    \includegraphics[width=1\linewidth]{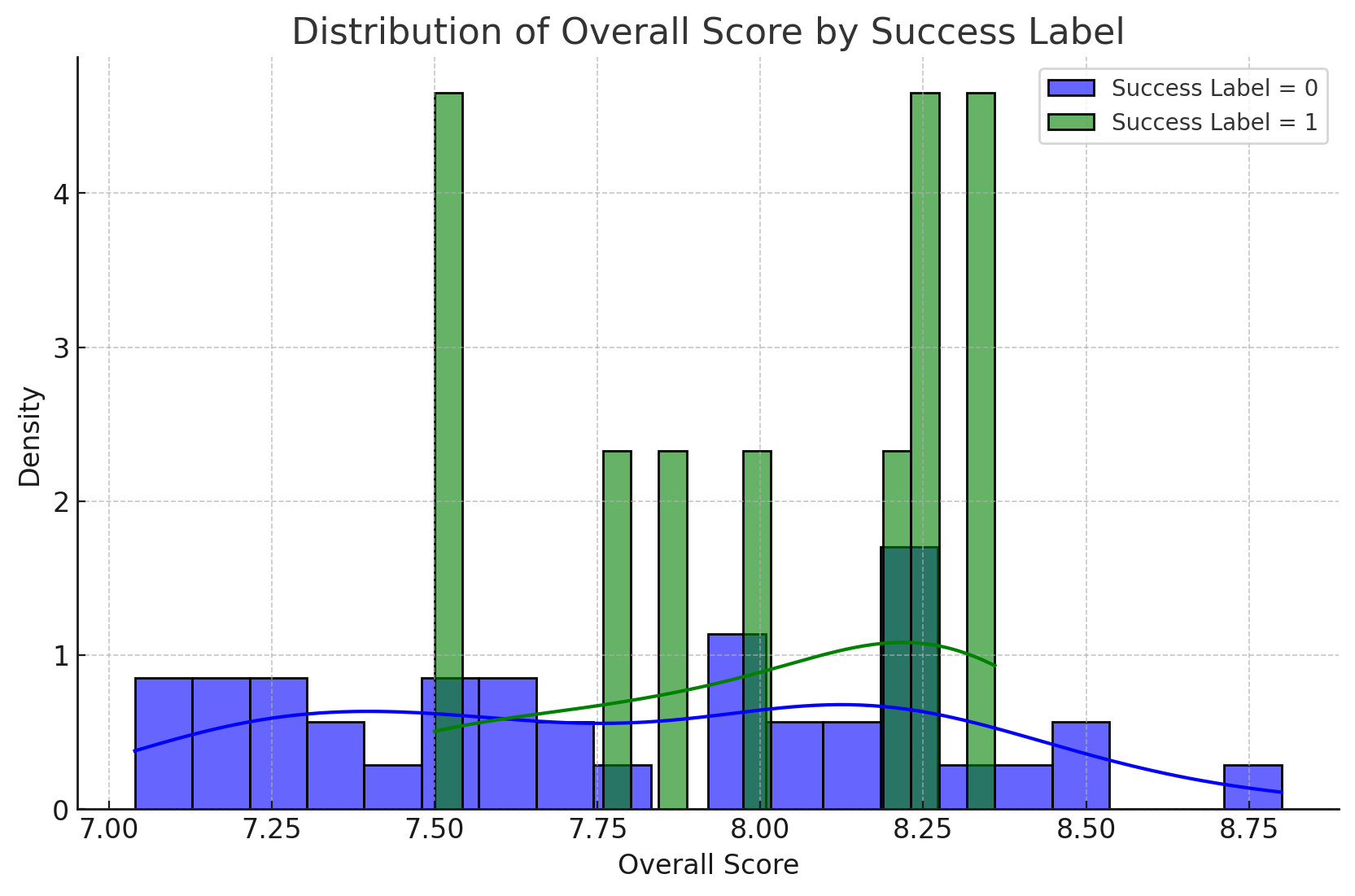}
    \caption{Distribution of Overall Score by Success Label}
    \label{fig:enter-label}
\end{figure}

\paragraph{Over-prediction Bias in LLMs}
Our results indicate that baseline LLMs, such as GPT-4o, exhibit a pronounced over-prediction bias when evaluating startup success. Although GPT-4o achieves perfect recall (100\%), its precision is extremely low (21.28\%), leading to an overall accuracy of only 26.00\% and a high false positive rate. These findings suggest that, in their default state, LLMs tend to overestimate the likelihood of startup success—likely due to overtrusting founder claims—thereby compromising their reliability in practical, real-world VC sourcing.

\paragraph{Performance Improvement With SSFF}
In contrast, the SSFF framework significantly enhances predictive performance and reliability. The SSFF-4o variant improves categorical prediction accuracy to 34\% while maintaining perfect recall, and its SSFF-4o-mini variant further elevates performance, achieving a final outcome accuracy of 50.00\% with robust recall (90.00\%) and precision (27.27\%). Additionally, our analysis of founder segmentation reveals that successful startups have an average segmentation score of 3.8, compared to 3.275 for unsuccessful ones. These improvements stem from SSFF’s integration of multiple perspectives, timely external information retrieval, and reliable predictions from its dedicated prediction blocks. The 108.3\% relative improvement over GPT-4o-mini and a 30.8\% relative improvement over GPT-4o validates the adoption of a multi-agent framework as a promising approach to achieving a more effective and interpretable decision-support system for startup evaluations.

\begin{figure}
    \centering
    \includegraphics[width=1\linewidth]{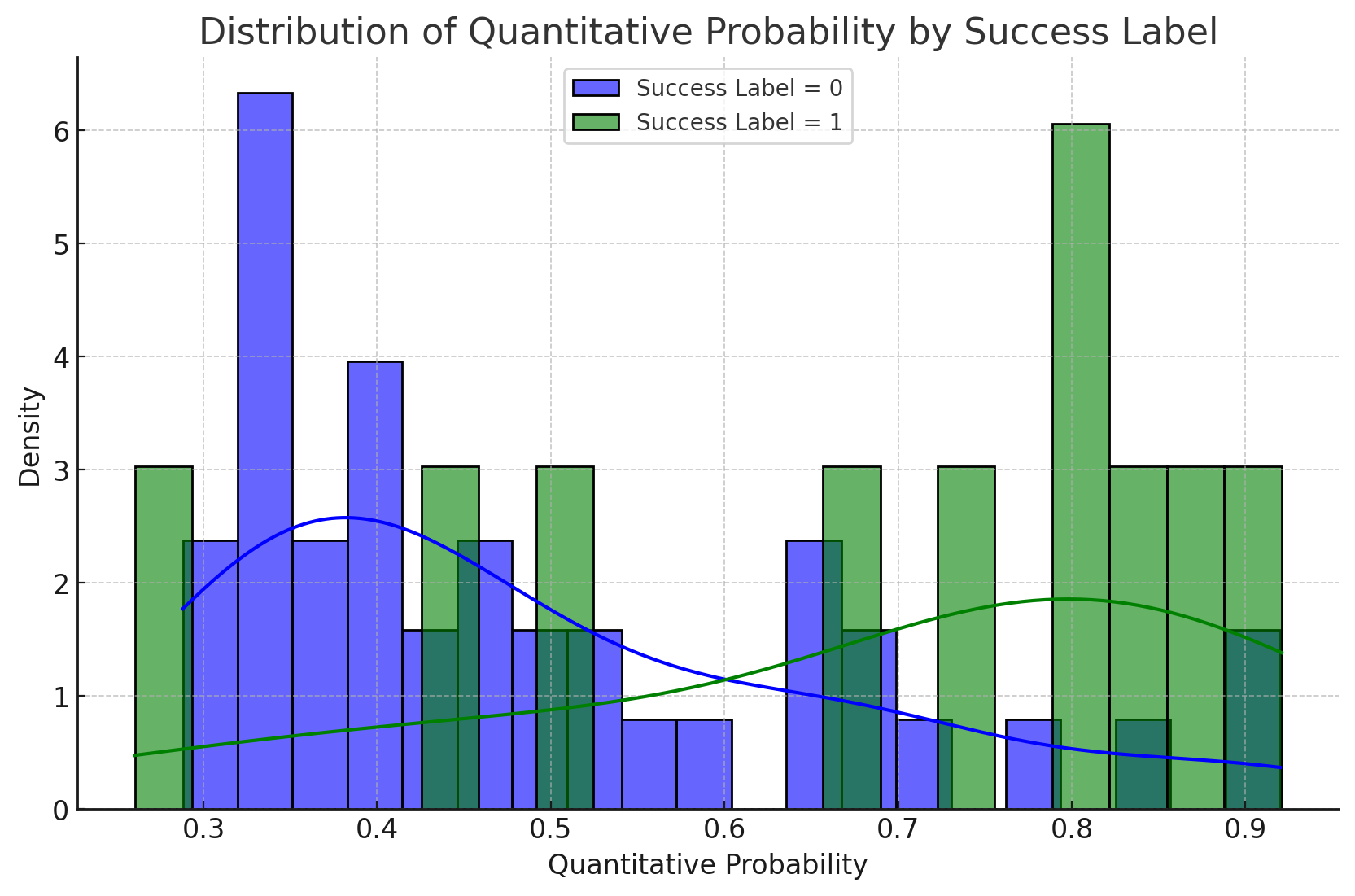}
    \caption{Distribution of Level of Confidence by Model Output}
    \label{fig:enter-label}
\end{figure}

\begin{table}[H]
\centering
\caption{Performance Metrics Comparison}
\label{tab:perf_metrics}
\small
\begin{tabular}{@{}lcccc@{}}
\toprule
\textbf{Model}               & \textbf{Accuracy}  \\ 
\midrule
Baseline (GPT-4o)            & 26.00\%                            \\
Baseline (GPT-4o-mini)       & 24.00\%                        \\
SSFF-4o                      & \textbf{34.00\%}          \\ 
SSFF-4o-mini                 & \textbf{50.00\%}         \\ 
\bottomrule
\end{tabular}
\vspace{-3mm}
\end{table}

\paragraph{Additional Observations}
Beyond the clear performance improvements achieved by SSFF, our research reveals intriguing interactions between LLM predictions and the startup evaluation process. For instance, the variance in overall scores is noticeably larger for non-successful startups compared to successful ones, indicating that predictions for failing startups are more dispersed. Additionally, the quantitative probabilities—reflecting the LLMs’ confidence levels—exhibit an inverse trend, with lower variance observed in “Hold” recommendations than in “Invest” decisions. These observations offer valuable insights into the underlying behavior of LLMs in startup success prediction, a key focus of the SSFF framework. Further investigations are attached in the appendix.

\subsection{Qualitative Evaluations}

Human expert evaluations highlight that SSFF offers significantly greater transparency and reliability compared to zero-shot GPT, enriching its analysis with external information and providing structured, dimension-based scoring. This design ensures more comprehensive and specific evaluations, instilling confidence in users by avoiding the vague and generalized observations typical of Vanilla GPT. Instead, SSFF delivers actionable insights and rigorous risk assessments, excelling in scenarios requiring detailed analysis. Its modular and adaptable framework further supports continual improvement, making SSFF a highly promising tool for both research advancements and commercial applications.

\section{Conclusion and Future Work}

\paragraph{Conclusion}
In this study, we explored the application of large language models (LLMs) in predicting startup success. Our work makes several key contributions:

\begin{enumerate}
\item \textbf{Identification of Over-Prediction Bias:} Our analysis revealed that LLMs tend to over-predict startup success, leading to high false positive rates.
\item \textbf{Effective Founder Segmentation:} We demonstrated that incorporating detailed founder segmentation significantly improves prediction accuracy by quantitatively validating its impact on startup outcomes.
\item \textbf{Novel SSFF Framework:} We introduced the \textbf{Startup Success Forecasting Framework (SSFF)}, a hybrid multi-agent system that integrates conventional machine learning models with LLMs. SSFF leverages a custom-trained model and real-time external market insights to achieve marked improvements over baseline LLMs, enhancing both reliability and transparency.
\item \textbf{Insights into LLM Behaviors:} We defined various metrics to analyze LLM behavior—such as confidence levels and variance in predictions—and uncovered intriguing patterns that further our understanding of LLM performance in data-constrained startup evaluations.
\end{enumerate}

This work not only highlights the limitations of current LLMs in startup evaluation but also paves the way for future research in developing advanced, interpretable decision-support systems for venture capital.

\paragraph{Limitations}
Our work has several limitations. First, our evaluation relies on a relatively small, stratified sample of 50 startups, which may not fully capture the diversity of real-world startup conditions. Second, while SSFF significantly improves upon baseline LLM performance, the absolute accuracy remains modest, and the predictions are still susceptible to inherent LLM biases—such as over-prediction of startup success. Third, our analysis is based on minimal input data (e.g., brief startup descriptions and limited founder information), which may not encompass all the nuanced factors influencing startup outcomes. Finally, although we introduce multiple evaluation metrics and a multi-agent approach, further validation through larger-scale experiments and human expert evaluations is necessary to confirm the robustness and practical applicability of our framework.

\paragraph{Future Work} Future work will follow two primary directions: deeper investigation of LLM behaviors and continued refinement of the SSFF framework. We plan to expand our baseline comparisons by incorporating additional models such as \textbf{Llama}, \textbf{Claude}, \textbf{Qwen}, and others. Several key metrics—including \textbf{FIFS distributions}, \textbf{founder-level segmentation differences}, and \textbf{confidence level variations}—warrant detailed analysis (a preliminary showcase is provided in the Appendix). Furthermore, increasing the sample size will enable more rigorous evaluations beyond our initial 50 startups. We also intend to explore more advanced models, such as \textbf{OpenAI’s O-series} and \textbf{DeepSeek’s V3 and R1 models}, to further enhance training and evaluation. To better demonstrate SSFF’s transparency and reasoning capabilities, we advocate for the involvement of human evaluators and plan to build a dataset pairing SSFF outputs with human annotations or preference rankings. Finally, further investigation into the over-prediction bias is necessary to develop effective mitigation strategies.

\appendix

\section{Appendix}

Comprehensive details on our methodologies, data processing, model architectures, additional experimental results, and code are provided as open-source resources in the appendix. Here, we include key information for transparency and reproducibility.

\subsection{Founders Segmentation}

\subsubsection{Segmentation Criteria}

The segmentation of founders into levels L1 through L5 is achieved through a supervised, hybrid approach that leverages both Large Language Models (LLMs) and manual expert review, drawing on years of venture capital expertise.

\begin{itemize}
\item \textbf{Level 1 (L1):} Founders with negligible experience—typically recent graduates, dropouts, or individuals largely disconnected from tech circles.
\item \textbf{Level 2 (L2):} Founders with limited experience (usually fewer than 10 years) who have worked at reputable companies (e.g., Google or McKinsey) or are accelerator graduates.
\item \textbf{Level 3 (L3):} First--time entrepreneurs with 10–-15 years of technical and management experience, often including individuals with advanced degrees (e.g., PhDs) or backgrounds from top-tier institutions.
\item \textbf{Level 4 (L4):} Repeat entrepreneurs who have successfully exited previous ventures via small to medium-sized exits, or professionals with high-level executive experience at notable technology companies.
\item \textbf{Level 5 (L5):} Elite founders who have built companies achieving substantial milestones—such as \$100M+ ARR, IPOs, or sales exceeding USD 500M—considered the most successful and influential in the ecosystem.
\end{itemize}

\subsubsection{Founders Segmentation Prompt} 

This section contains the complete prompt template used to categorize founders into segmentation levels (L1–-L5). The prompt instructs the LLM to evaluate a founder’s background (including education, work experience, leadership roles, and achievements) and assign a segmentation level accordingly.

\textbf{Example Prompt:}
\begin{lstlisting}[basicstyle=\ttfamily\small, breaklines=true, frame=single]
You are an analyst. Your task is to output one of the options: [L1, L2, L3, L4, L5]. Do not output anything else.

Think step by step and consider the following criteria:
- L5: Entrepreneur who has built a $100M+ ARR business, taken a company public, or achieved a sale exceeding $500M.
- L4: Entrepreneur with a small to medium-size exit or who has held a high-level executive role at a notable technology company.
- L3: First time entrepreneur with 10 to 15 years of technical and management experience, often holding advanced degrees or coming from top-tier institutions.
- L2: Entrepreneur with a few years of experience or an accelerator graduate.
- L1: Entrepreneur with negligible experience (e.g., recent graduate or dropout) but with potential.

Based on the founder's LinkedIn profile information below, determine the appropriate segmentation level:
{founder_info}
\end{lstlisting}

(Additional prompt versions and iterative refinements are also documented in our version-controlled repository.)

\subsection{Analyst Block Implementation Details}

\subsubsection{Prompts}

\textbf{Product Agent}

\begin{lstlisting}[basicstyle=\ttfamily\small, breaklines=true, frame=single]
You are a professional product analyst in a VC firm evaluating a potential investment opportunity.

Company Information:
{startup_info}

Product Information:
{product_info}

Product Research Report:
{external_knowledge}

Based on this comprehensive product research and the initial data, please provide:
1.Technical Innovation Analysis:
-How innovative is the technology?
-Is it feasible to implement?
-What are the technical risks?

2.Feature Set Evaluation:
-How complete is the product's feature set?
-How does it compare to competitors?
-What are the key differentiators?

3.Implementation Assessment:
-What are the main technical challenges?
-How realistic is the development timeline?
-What resources are required?

4.Market Readiness:
-Is the product ready for its target market?
-What further development is needed?
-How strong is the product-market fit?

Please reference specific data points from the product research report in your analysis, and conclude with:
-Product potential score (1-10)
-Innovation score (1-10)
-Market fit score (1-10)
\end{lstlisting}

\textbf{Market Agent}

\begin{lstlisting}[basicstyle=\ttfamily\small, breaklines=true, frame=single]
You are a professional agent in a VC firm to analyze a company. Your task is to analyze the company here. Context: {startup_info}

Your focus is on the market side. What is the market? Is the market big enough? Is now the good timing? Will there be a good product-market-fit? 
        
Specifically here are some relevant market information: {market_info}. 

Your intern has researched more around the following topic for you as context {keywords}.

The research result: {external_knowledge}

Provide a comprehensive analysis including market size, growth rate, competition, and key trends. Analyze step by step to formulate your comprehensive analysis to answer the questions proposed above.

Also conclude with a market viability score from 1 to 10. 
\end{lstlisting}

\textbf{Founders Agent}

\begin{lstlisting}[basicstyle=\ttfamily\small, breaklines=true, frame=single]
As a highly qualified analyst specializing in startup founder assessment, evaluate the founding team based on the provided information.
        Consider the founders' educational background, industry experience, leadership capabilities, and their ability to align and execute on the company's vision.
        Provide a competency score, key strengths, and potential challenges. Please write in great details.
\end{lstlisting}

Sample input is here. 

\begin{lstlisting}[basicstyle=\ttfamily\small, breaklines=true, frame=single]
startup-info = {
    "founder_backgrounds": "MBA from Stanford, 5 years at Google as Product Manager",
    "track_records": "Successfully launched two products at Google, one reaching 1M users",
    "leadership_skills": "Led a team of 10 engineers and designers",
    "vision_alignment": "Strong passion for AI and its applications in healthcare",
    "description": "AI-powered health monitoring wearable device"
}
\end{lstlisting}

\paragraph{Integration Agent}

We have two prompts attached here, one for integration decision and another for quantitative decision (to separate out reasoning from sub-reports).

\textbf{Integration Analysis Prompt}

\begin{lstlisting}[basicstyle=\ttfamily\small, breaklines=true, frame=single]
Imagine you are the chief analyst at a venture capital firm, tasked with integrating the analyses of multiple specialized teams to provide a comprehensive investment insight. Your output should be structured with detailed scores and justifications:
        
As the chief analyst, you should stay critical of the company and listen carefully to what your colleagues say. You are also assisted by statistical models trained by your firm. You should not be over confident (or over-critical) for a firm and should rely on your strength of reasoning. 
Many startups present themselves with good words but the truth is that few will be successful. It is your task to find those that have the potential to be successful and give your recommendations. 

Example 1:
Market Viability: 8.23/10 - The market is on the cusp of a regulatory shift that could open up new demand channels, supported by consumer trends favoring sustainability. Despite the overall growth, regulatory uncertainty poses a potential risk.
Product Viability: 7.36/10 - The product introduces an innovative use of AI in renewable energy management, which is patent-pending. However, it faces competition from established players with deeper market penetration and brand recognition.
Founder Competency: 9.1/10 - The founding team comprises industry veterans with prior successful exits and a strong network in the energy sector. Their track record includes scaling similar startups and navigating complex regulatory landscapes.

Recommendation: Invest. The team's deep industry expertise and innovative product position it well to capitalize on the market's regulatory changes. Although competition is stiff, the founders' experience and network provide a competitive edge crucial for market adoption and navigating potential regulatory hurdles.

Example 2:
Market Viability: 5.31/10 - The market for wearable tech is saturated, with slow growth projections. However, there exists a niche but growing interest in wearables for pet health.
Product Viability: 6.5/10 - The startup's product offers real-time health monitoring for pets, a feature not widely available in the current market. Yet, the product faces challenges with high production costs and consumer skepticism about the necessity of such a device.
Founder Competency: 6.39/10 - The founding team includes passionate pet lovers with backgrounds in veterinary science and tech development. While they possess the technical skills and passion for the project, their lack of business and scaling experience is a concern.

Recommendation: Hold. The unique product offering taps into an emerging market niche, presenting a potential opportunity. However, the combination of a saturated broader market, challenges in justifying the product's value to consumers, and the team's limited experience in business management suggests waiting for clearer signs of product-market fit and strategic direction.

Now, analyze the following:

Market Viability: {market_info}
Product Viability: {product_info}
Founder Competency: {founder_info}
Founder-Idea Fit: {founder_idea_fit}
Founder Segmentation: {founder_segmentation}
Random Forest Prediction: {rf_prediction}

Some context here for the scores: 
1. Founder-Idea-Fit ranges from -1 to 1, a stronger number signifies a better fit.
2. Founder Segmentation outcomes range from L1 to L5, with L5 being the most "competent" founders, and L1 otherwise.
3. Random Forest Prediction predicts the expected outcome purely based on a statistical model, with an accuracy of around 65%.

Provide an overall investment recommendation based on these inputs. State whether you would advise 'Invest' or 'Hold', including a comprehensive rationale for your decision. Consider all provided predictions and analyses, but do not over-rely on any single prediction.
\end{lstlisting}

\textbf{QuantDecision Prompt}

\begin{lstlisting}[basicstyle=\ttfamily\small, breaklines=true, frame=single]
You are a final decision-maker. Think step by step. You need to consider all the quant metrics and makde a decision.

You are now given Founder Segmentation. With L5 very likely to succeed and L1 least likely. You are also given the Founder-Idea Fit Score, with 1 being most fit and -1 being least fit. You are also given the result of prediction model (which should not be your main evidence because it may not be very accurate).

This table summarises the implications of the Level Segmentation:

Founder Level & Success & Failure & Success Rate & X-Time Better than L1 \\
\midrule
L1 & 24 & 75 & 24.24\% & 1 \\
L2 & 83 & 223 & 27.12\% & 1.12 \\
L3 & 287 & 445 & 39.21\% & 1.62 \\
L4 & 514 & 249 & 67.37\% & 2.78 \\
L5 & 93 & 8 & 92.08\% & 3.79 \\

Regarding the Founder-Idea-Fit Score. Relevant context are provided here: 
The previous sections show the strong correlation between founder's segmentation level and startup's outcome, as L5 founders are more than three times likely to succeed than L1 founders. However, looking into the data, one could also see that there are L5 founders who did not succeed, and there are L1 founders who succeeded. To account for these scenarios, we investigate the fit between founders and their ideas.

To assess quantitatively, we propose a metric called Founder-Idea Fit Score (FIFS). The Founder-Idea Fit Score quantitatively assesses the compatibility between a founder's experience level and the success of their startup idea. Given the revised Preliminary Fit Score ($PFS$) defined as:
\[PFS(F, O) = (6 - F) \times O - F \times (1 - O)\]
where $F$ represents the founder's level ($1$ to $5$) and $O$ is the outcome ($1$ for success, $0$ for failure), we aim to normalize this score to a range of $[-1, 1]$ to facilitate interpretation.

To achieve this, we note that the minimum $PFS$ value is $-5$ (for a level $5$ founder who fails), and the maximum value is $5$ (for a level $1$ founder who succeeds). The normalization formula to scale $PFS$ to $[-1, 1]$ is:
\[Normalized\;PFS = \frac{PFS}{5}\]

Now use all of these information, produce a string of the predicted outcome and probability, with one line of reasoning. 

Your response should be in the following format:
{
  "outcome": "<Successful or Unsuccessful>",
  "probability": <probability as a float between 0 and 1>,
  "reasoning": "<One-line reasoning for the decision>"
}

You will also receive a categorical prediction outcome of the prediction model (which should not be your main evidence because it may not be very accurate, just around 65% accuracy).

Ensure that your response is a valid JSON object and includes all the fields mentioned above.

f"You are provided with the categorical prediction outcome of {rf_prediction}, Founder Segmentation of {Founder_Segmentation}, Founder-Idea Fit of {Founder_Idea_Fit}."

\end{lstlisting}

\subsection{Prediction Block Implementation Details}

\subsubsection{VC Scout Agent}

\begin{lstlisting}[basicstyle=\ttfamily\small, breaklines=true, frame=single]
As an analyst specializing in startup evaluation, categorize the given startup based on the following criteria.
Provide a categorical response for each of the following questions based on the startup information provided.
Use ONLY the specified categorical responses for each field. Do not use any other responses.

1. Industry Growth: [Yes/No/N/A]
2. Market Size: [Small/Medium/Large/N/A]
3. Development Pace: [Slower/Same/Faster/N/A]
4. Market Adaptability: [Not Adaptable/Somewhat Adaptable/Very Adaptable/N/A]
5. Execution Capabilities: [Poor/Average/Excellent/N/A]
6. Funding Amount: [Below Average/Average/Above Average/N/A]
7. Valuation Change: [Decreased/Remained Stable/Increased/N/A]
8. Investor Backing: [Unknown/Recognized/Highly Regarded/N/A]
9. Reviews and Testimonials: [Negative/Mixed/Positive/N/A]
10. Product-Market Fit: [Weak/Moderate/Strong/N/A]
11. Sentiment Analysis: [Negative/Neutral/Positive/N/A]
12. Innovation Mentions: [Rarely/Sometimes/Often/N/A]
13. Cutting-Edge Technology: [No/Mentioned/Emphasized/N/A]
14. Timing: [Too Early/Just Right/Too Late/N/A]

Provide your analysis in a JSON format that matches the StartupCategorization schema.
If you cannot determine a category based on the given information, use 'N/A'.
Do not include any explanations or additional text outside of the JSON structure.

Startup Information:{startup_info}
\end{lstlisting}

\subsubsection{LLM-based RF Model}

Mapping is attached here. 

\begin{lstlisting}[basicstyle=\ttfamily\small, breaklines=true, frame=single]
"industry_growth": ["No", "N/A", "Yes"],
"market_size": ["Small", "Medium", "Large", "N/A"],
"development_pace": ["Slower", "Same", "Faster", "N/A"],
"market_adaptability": ["Not Adaptable", "Somewhat Adaptable", "Very Adaptable", "N/A"],
"execution_capabilities": ["Poor", "Average", "Excellent", "N/A"],
"funding_amount": ["Below Average", "Average", "Above Average", "N/A"],
"valuation_change": ["Decreased", "Remained Stable", "Increased", "N/A"],
"investor_backing": ["Unknown", "Recognized", "Highly Regarded", "N/A"],
"reviews_testimonials": ["Negative", "Mixed", "Positive", "N/A"],
"product_market_fit": ["Weak", "Moderate", "Strong", "N/A"],
"sentiment_analysis": ["Negative", "Neutral", "Positive", "N/A"],
"innovation_mentions": ["Rarely", "Sometimes", "Often", "N/A"],
"cutting_edge_technology": ["No", "Mentioned", "Emphasized", "N/A"],
"timing": ["Too Early", "Just Right", "Too Late", "N/A"]
\end{lstlisting}

\begin{table}[ht]
\centering
\begin{tabular}{cc|c|c|}
\cline{3-4}
& & \multicolumn{2}{ c| }{Predicted Class} \\ 
\cline{3-4}
& & 0 & 1 \\ 
\cline{1-4}
\multicolumn{1}{ |c  }{\multirow{2}{*}{Actual Class}} &
\multicolumn{1}{ |c| }{0} & 99 & 38 \\ 
\cline{2-4}
\multicolumn{1}{ |c  }{}                        &
\multicolumn{1}{ |c| }{1} & 27 & 115 \\ 
\cline{1-4}
\end{tabular}
\caption{Confusion matrix for LLM-based random forest model.}
\label{tab:confusion_matrix}
\end{table}

\subsection{External Knowledge Block – Detailed Technical Supplement}

\textbf{Market Research Synthesis Prompt}

\begin{lstlisting}[basicstyle=\ttfamily\small, breaklines=true, frame=single]
You are a market research analyst. Synthesize the search results focusing on quantitative data points:
    
- Market size (in USD)
- Growth rates (CAGR)
- Market share percentages
- Transaction volumes
- Customer acquisition costs
- Revenue metrics
 - Competitive landscape metrics
    
Format data points clearly and cite their time periods. If exact numbers aren't available, provide ranges based on available data. Prioritize numerical data over qualitative descriptions.
\end{lstlisting}

\begin{enumerate}
\item \textbf{External Knowledge Generation Details}
\begin{itemize}
\item \textbf{Keyword Generation:}
\begin{itemize}
\item Attach the complete prompt template used for generating keywords.
\item Provide examples of generated keywords for sample inputs, such as the “Company (redacted)” example.
\end{itemize}
\item \textbf{Content Retrieval:}
\begin{itemize}
\item Include code snippets that demonstrate how the SERP API is called.
\item Present sample raw search results (e.g., news articles, blog posts) before filtering.
\end{itemize}
\item \textbf{Data Filtering:}
\begin{itemize}
\item Detail the filtering process, including criteria for selecting high-quality content.
\item Provide examples of filtered content and a discussion of how it enhances data quality.
\end{itemize}
\end{itemize}
\item \textbf{Market Analysis Process and Comparative Evaluation}
\begin{itemize}
    \item \textbf{Parameter Variation (N=3 vs. N=10):}  
    \begin{itemize}
        \item Include a description and code examples of how the SERP API is configured with different values of $begin:math:text$N$end:math:text$ (e.g., $begin:math:text$N=3$end:math:text$ and $begin:math:text$N=10$end:math:text$).
        \item Attach a comparative table or graph showing the differences in data depth, structured insights, and timeliness between the two settings.
    \end{itemize}
    \item \textbf{Performance Comparison:}  
    \begin{itemize}
        \item Summarize the advantages observed when scaling from $begin:math:text$N=3$end:math:text$ to $begin:math:text$N=10$end:math:text$, such as improved granularity and relevance of insights.
    \end{itemize}
\end{itemize}

\item \textbf{Insight Synthesis Explanation}
\begin{itemize}
    \item \textbf{Prompt Templates:}  
    \begin{itemize}
        \item Attach the detailed prompts used to guide LLMs (e.g., GPT or Claude) for synthesizing the retrieved data into market reports.
    \end{itemize}
    \item \textbf{Report Generation Examples:}  
    \begin{itemize}
        \item Include sample outputs (market reports) generated by the LLM.
        \item Provide a discussion on how these outputs are integrated into the SSFF framework.
    \end{itemize}
\end{itemize}

\item \textbf{Case Study Example}
\begin{itemize}
    \item Present the full example input (the redacted company description) along with the corresponding generated keywords.
    \item Show how these keywords drive the subsequent search, filtering, and synthesis processes.
\end{itemize}
\end{enumerate}

\begin{figure}
    \centering
    \includegraphics[width=1\linewidth]{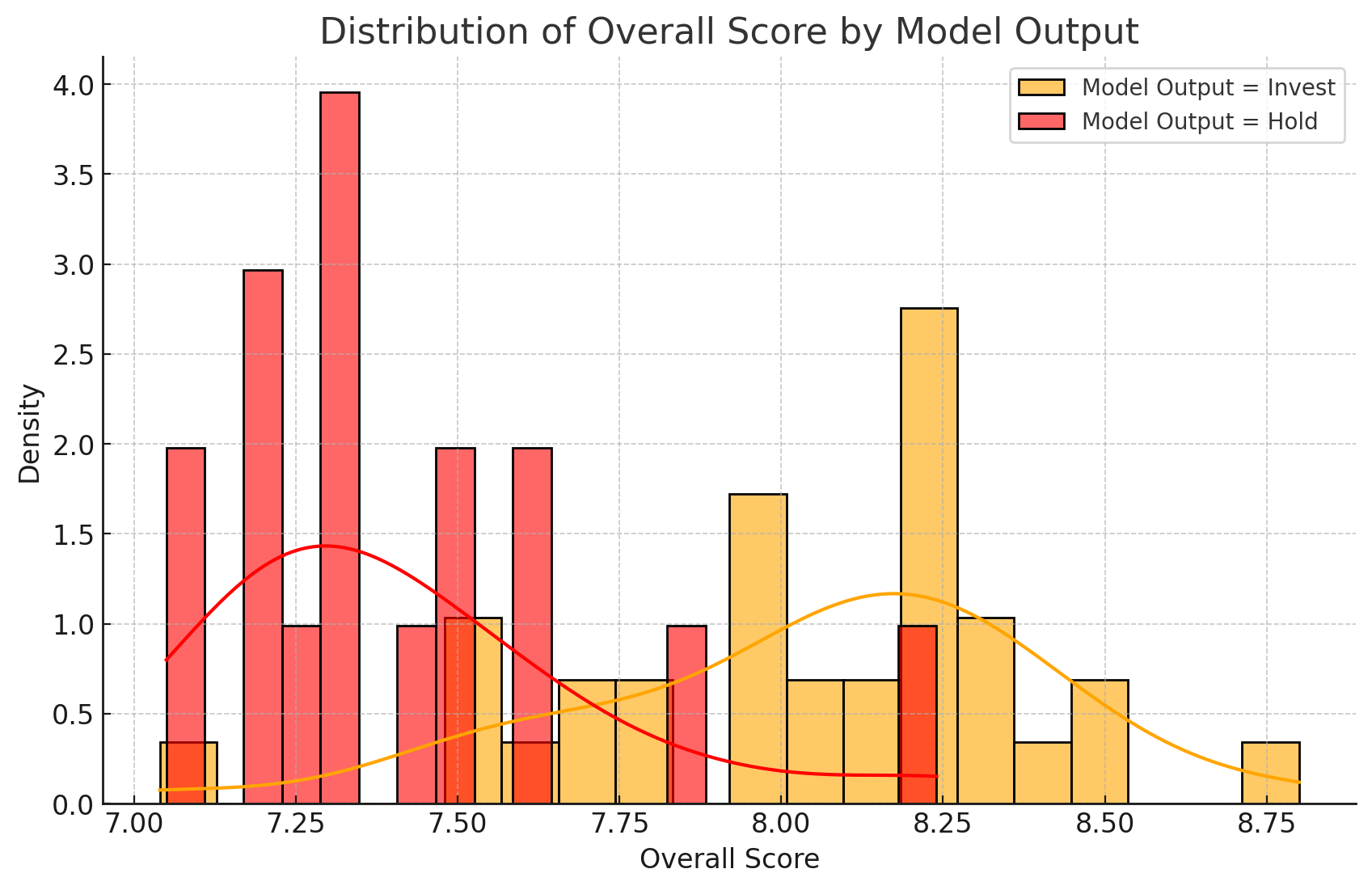}
    \caption{Distribution of Overall Score by Model Output}
    \label{fig:enter-label}
\end{figure}

\begin{figure}
    \centering
    \includegraphics[width=1\linewidth]{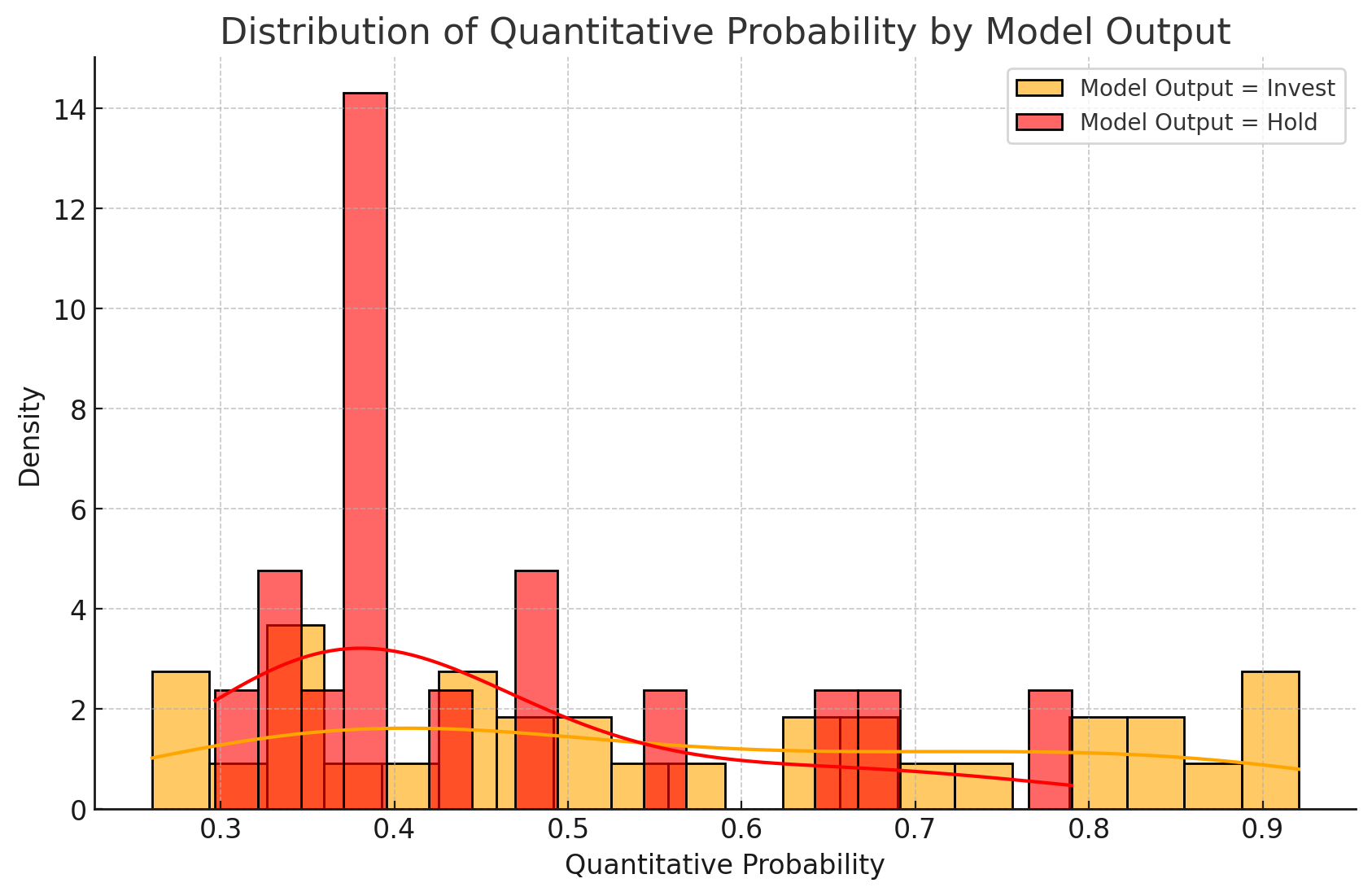}
    \caption{Distribution of Quantitative Probability (Level of Confidence) by Model Output}
    \label{fig:enter-label}
\end{figure}

Following the extraction of targeted information, the next step involves synthesizing this data into coherent and insightful market reports. This synthesis is accomplished through the use of LLMs (GPT, Claude, etc):

\begin{enumerate}
    \item \textbf{Prompt Design:} Crafting detailed prompts that guide the GPT model to analyze the collected data, considering the startup's context and the specifics of its market.
    \item \textbf{Report Generation:} Leveraging the GPT model's capabilities to integrate and interpret the data, generating comprehensive market reports that highlight key findings, opportunities, challenges, and trends.
    \item \textbf{Integration and Feedback:} Incorporating the synthesized market reports into SSFF's broader analytical framework, where they complement and enhance the insights provided by other blocks.
\end{enumerate}

For an example input like the one below, the corresponding keywords generated by SSFF for the search block is attached: 

\begin{quote}
"Company (redacted) aims to revolutionize China's \$2.5 billion college application consulting market by increasing access for over a million Chinese students aspiring to study abroad. As an AI-powered platform, Company (redacted) automates program selection, preparation guidance, and essay review using Large Language Models (LLM), the ANNOY Model, and an extensive database."
\end{quote}

\textbf{Keywords Generated:} Chinese Education Consulting Market, Growth, Trend, Size, Revenue.

\subsection{Additional Results Exploration}

\subsubsection{Analysis of Trends and Statistical Significance}
The visualizations reveal clear trends between metrics and outcomes. Figure 3-8 are based on SSFF-4o-mini version. For the \textbf{Success Label Groups}, higher \textit{Overall Scores} (mean difference visible in distributions) and \textit{Founder Competency Scores} are loosely associated with successful startups (Success Label = 1), while higher \textit{Quantitative Probabilities} exhibit a strong relationship with success. Similarly, for the \textbf{Model Output Groups}, Invest recommendations align with higher scores and probabilities, reflecting the model's tendency to identify promising cases, while Hold recommendations correlate with lower metrics. Statistical analysis supports these observations: the t-tests show no significant differences for \textit{Overall Scores} (t-statistic = -1.71, p = 0.103) and \textit{Founder Competency Scores} (t-statistic = 0.42, p = 0.683) between success labels, suggesting these metrics are not decisive predictors. However, \textit{Quantitative Probability} demonstrates a significant difference (t-statistic = -2.69, p = 0.019), highlighting its critical role in distinguishing successful startups and validating its importance in the model's predictions.

\begin{figure}
    \centering
    \includegraphics[width=1\linewidth]{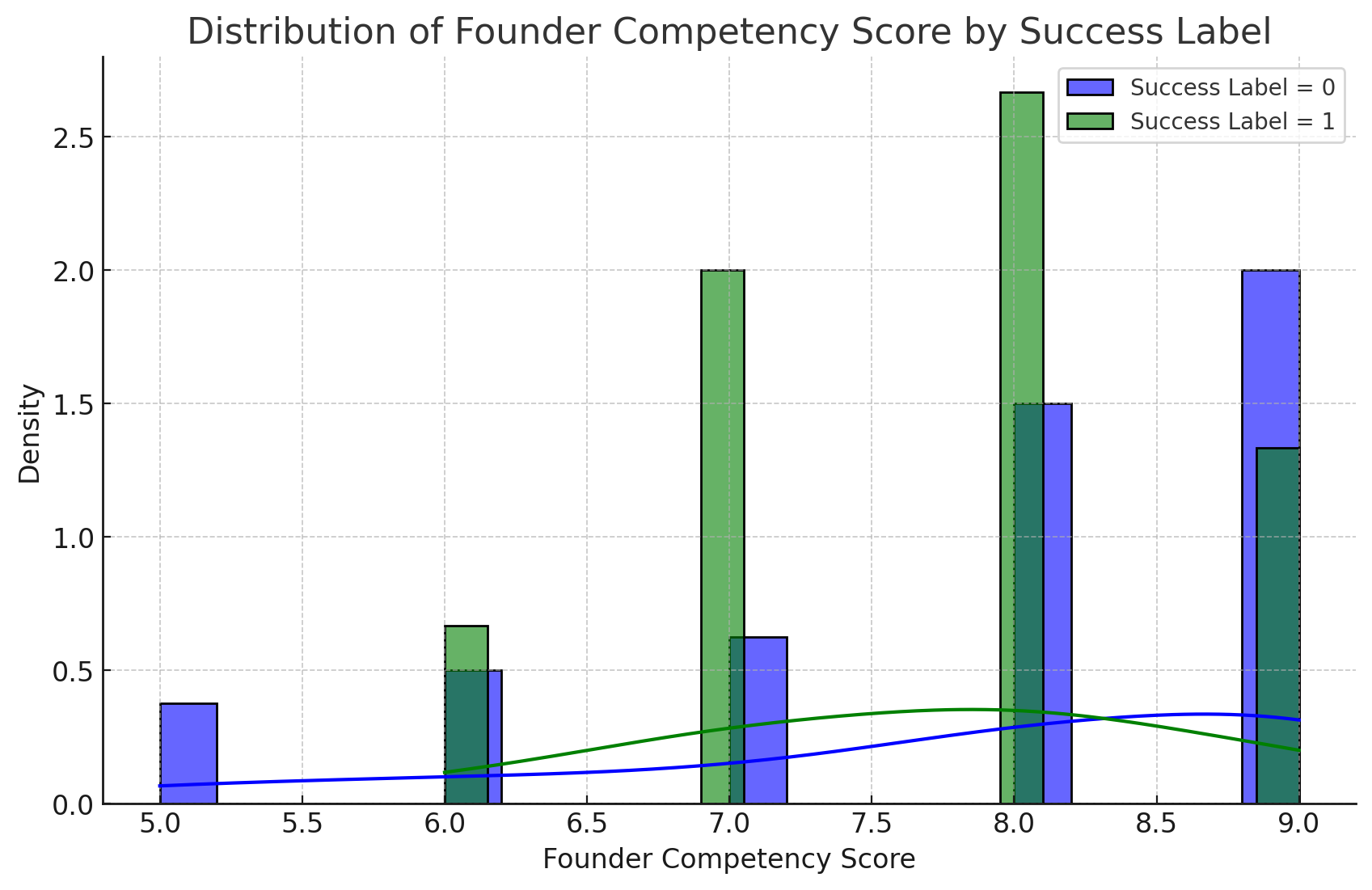}
    \caption{Distribution of Founder Competency Score by Success Label}
    \label{fig:enter-label}
\end{figure}

\subsubsection {Quantitative Metrics \& Impact on Prediction}

For the additional charts attached in the appendix, the analysis reveals distinct patterns in the relationship between founder competency score, idea fit score, and overall performance score with both ground truth success and predicted outcomes. Higher levels of these attributes are generally associated with successful cases, highlighting their importance as key indicators of success. However, deviations are observed when comparing predictions with actual outcomes, particularly in cases where intermediate probabilities lead to misclassification. This suggests that while the model effectively leverages these factors, there are biases or limitations in its ability to fully align with real-world success. 

\begin{figure}
    \centering
    \includegraphics[width=1\linewidth]{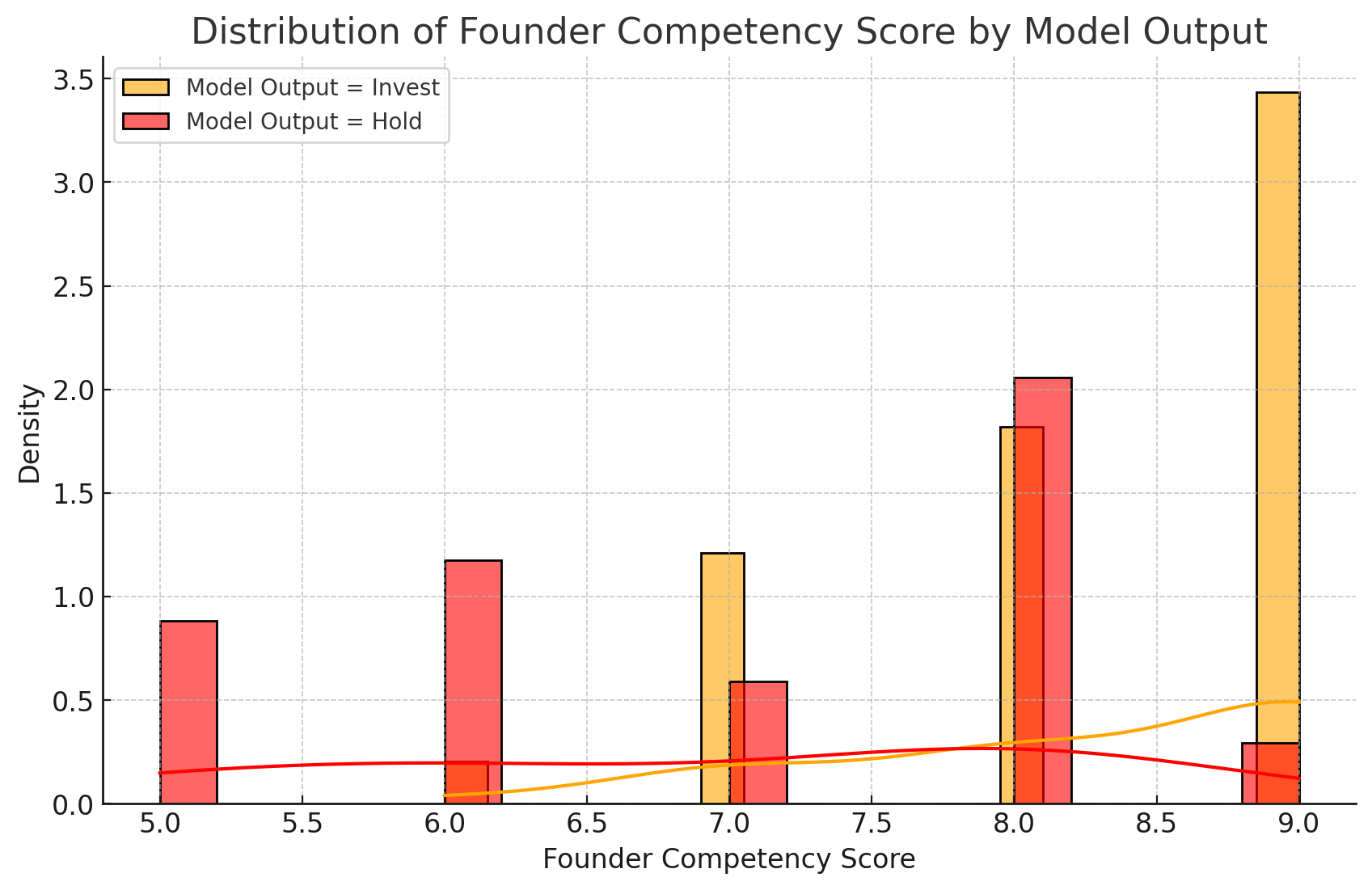}
    \caption{Distribution of Founder Competency Score by Model Output}
    \label{fig:enter-label}
\end{figure}

Another interesting observation is by comparing the scatter plot of founder competency score vs overall score in both clusters of ground truth and prediction. One could observe that the general pattern of success lies on the top right corner, but the prediction by SSFF has a higher range than the ground truth, resulting in many false positives. 

These findings emphasize the need for further refinement of the model to better account for the complexities of success prediction. Being able to showcase the quantitative metrics is a distinct characteristic in SSFF. 

\begin{figure}[t]
    \centering
    \includegraphics[width=1\linewidth]{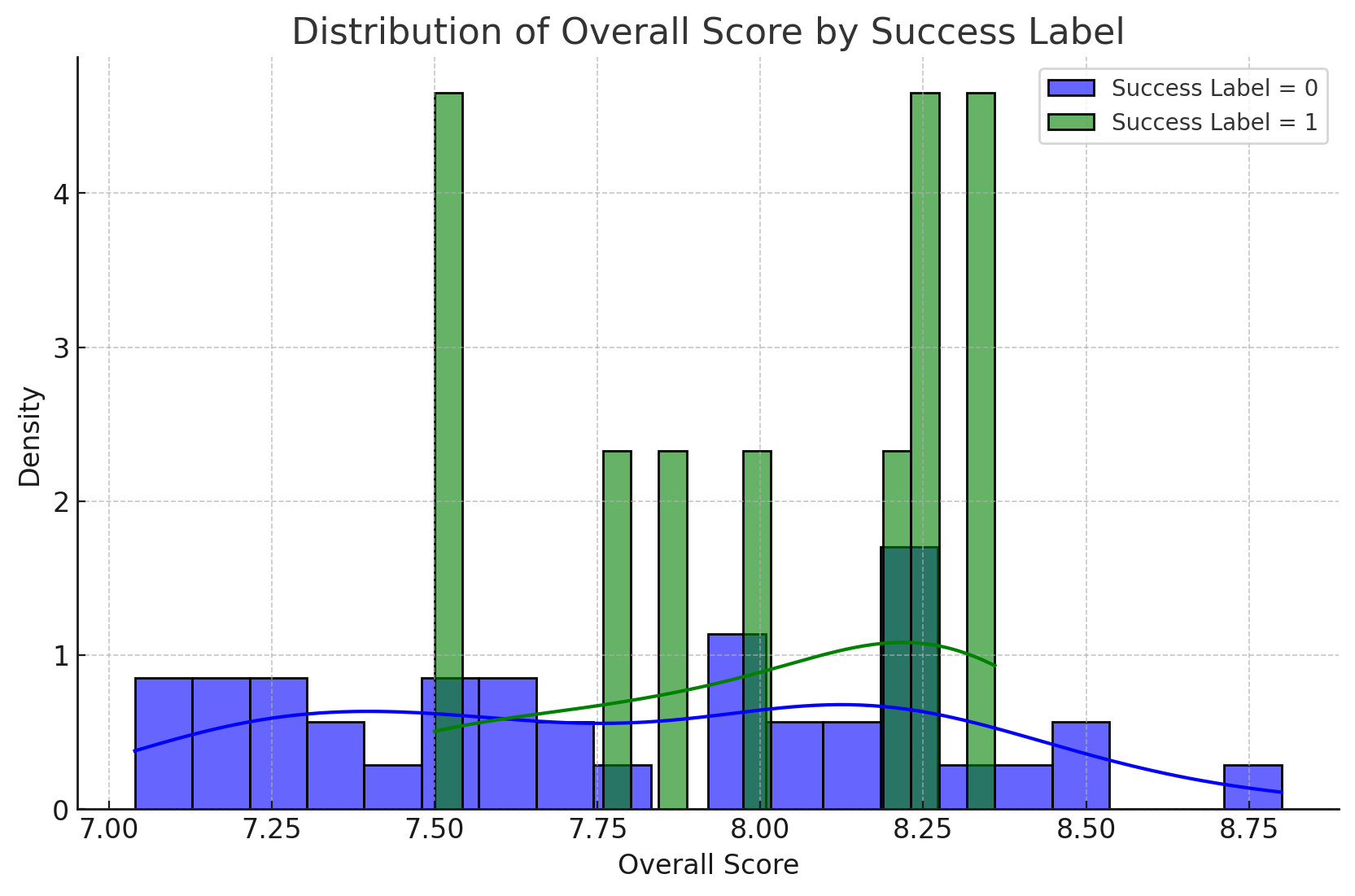}
    \caption{Histogram of Overall Score by Success (0 or 1)}
    \label{fig:enter-label}
\end{figure}

\begin{figure}
    \centering
    \includegraphics[width=1\linewidth]{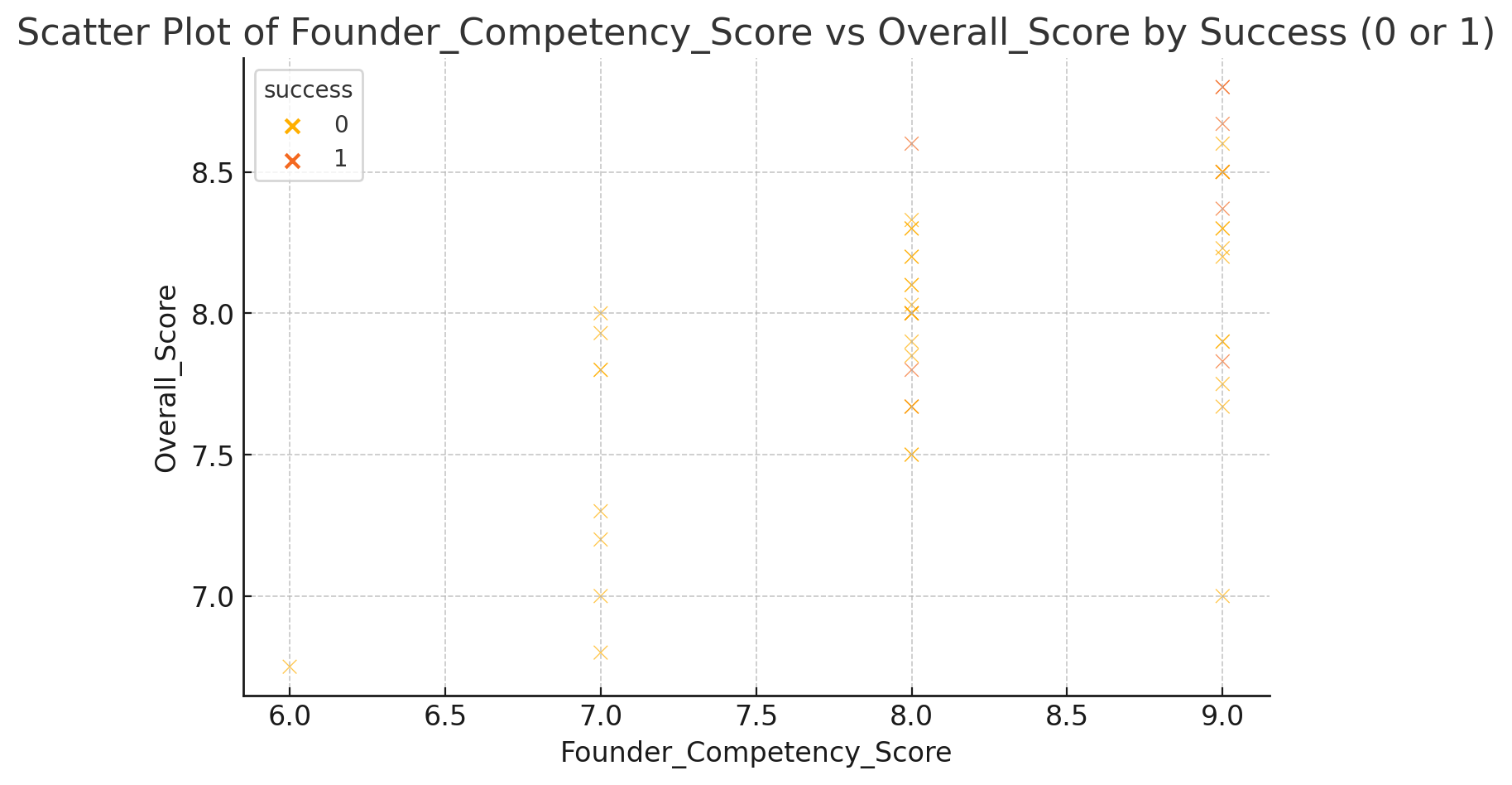}
    \caption{Scatter Plot of Founder Competency Score vs Overall Score by Success (0 or 1)}
    \label{fig:enter-label}
\end{figure}

\begin{figure}
    \centering
    \includegraphics[width=1\linewidth]{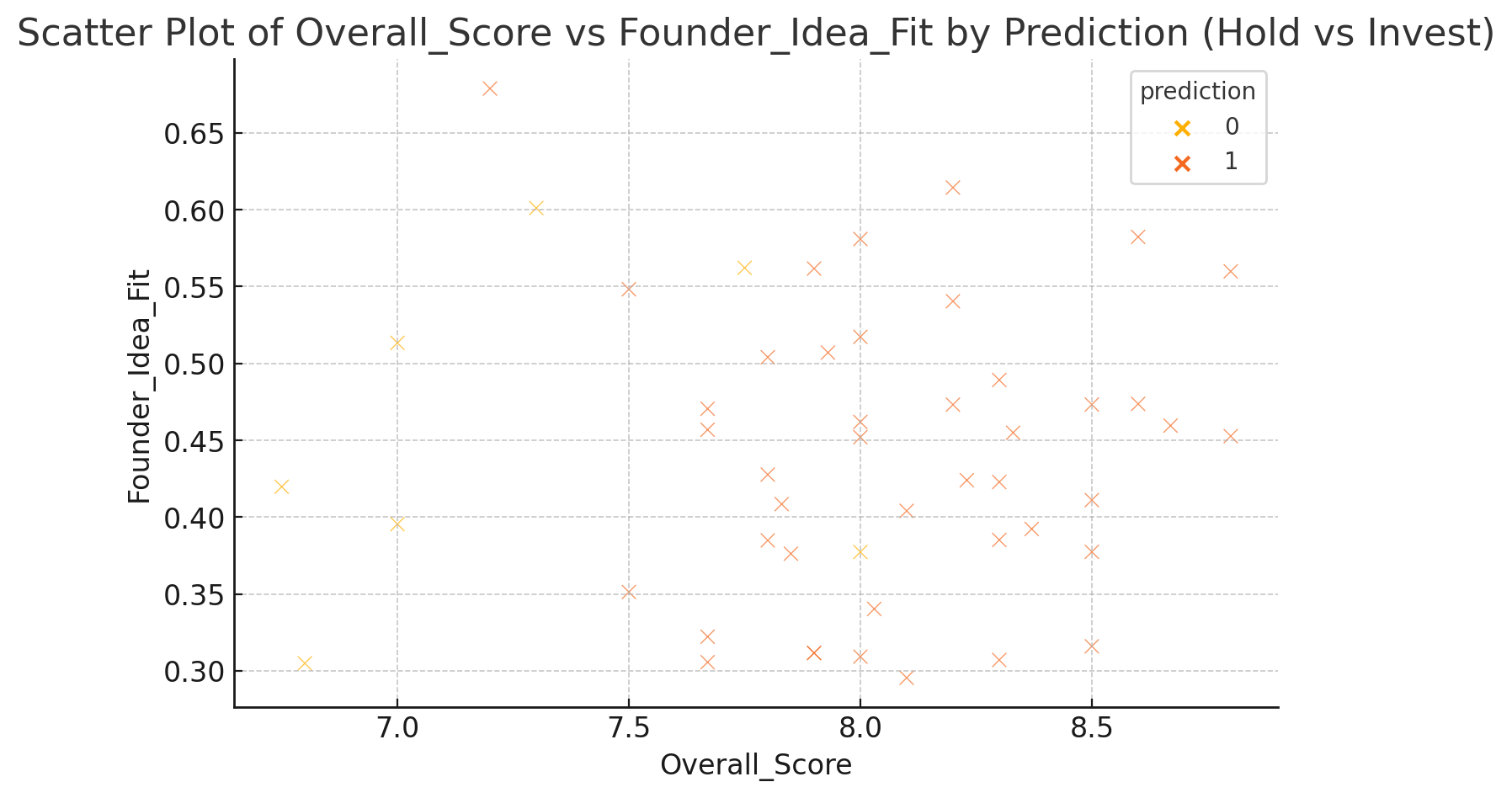}
    \caption{Scatter Plot of Overall Score vs Founder Idea Fit by Prediction (Hold vs Invest)}
    \label{fig:enter-label}
\end{figure}

\begin{figure}
    \centering
    \includegraphics[width=1\linewidth]{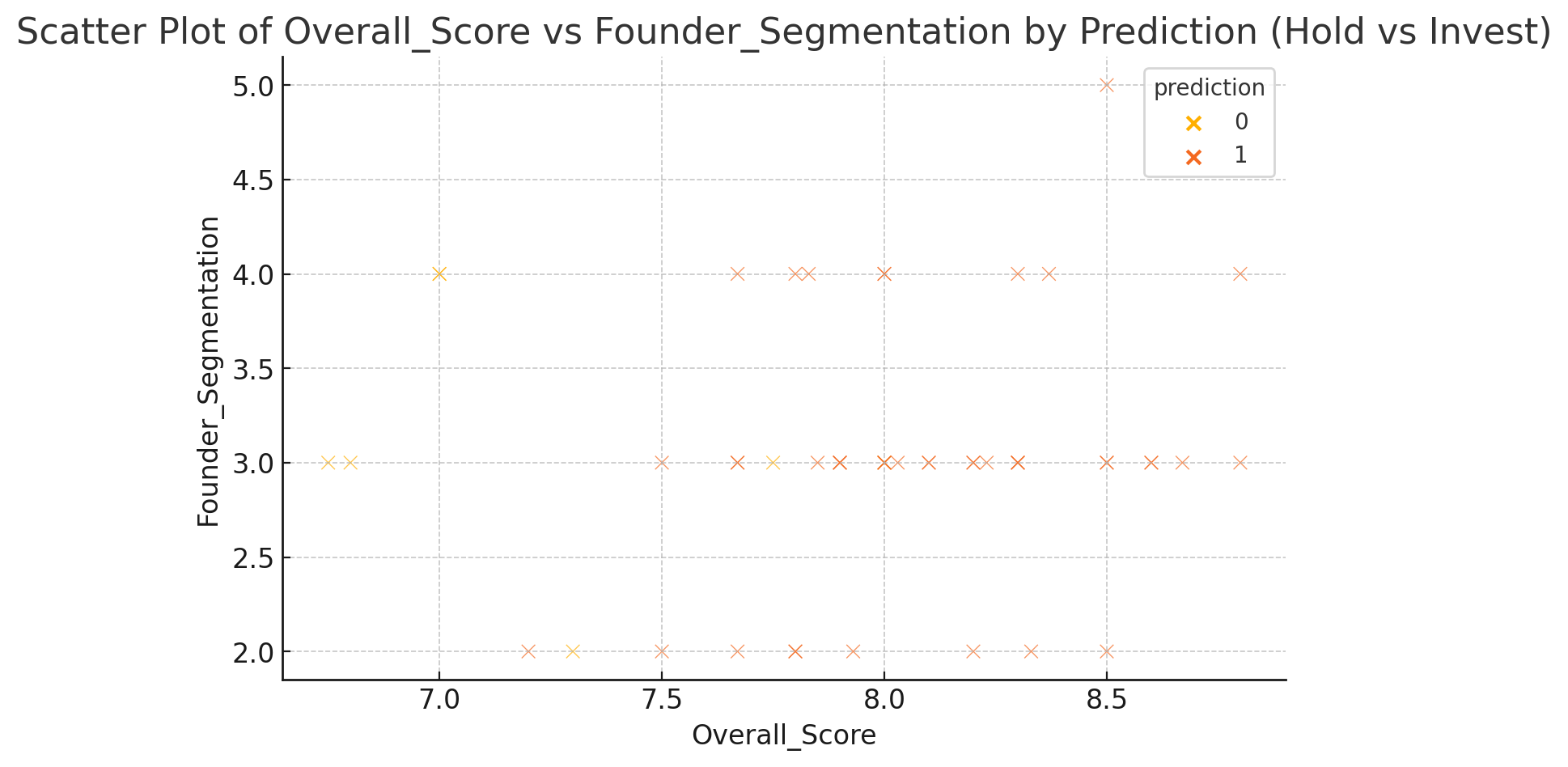}
    \caption{Scatter Plot of Overall Score vs Founder Segmentation by Prediction (Hold vs Invest)}
    \label{fig:enter-label}
    
\end{figure}

\begin{figure}
    \centering
    \includegraphics[width=1\linewidth]{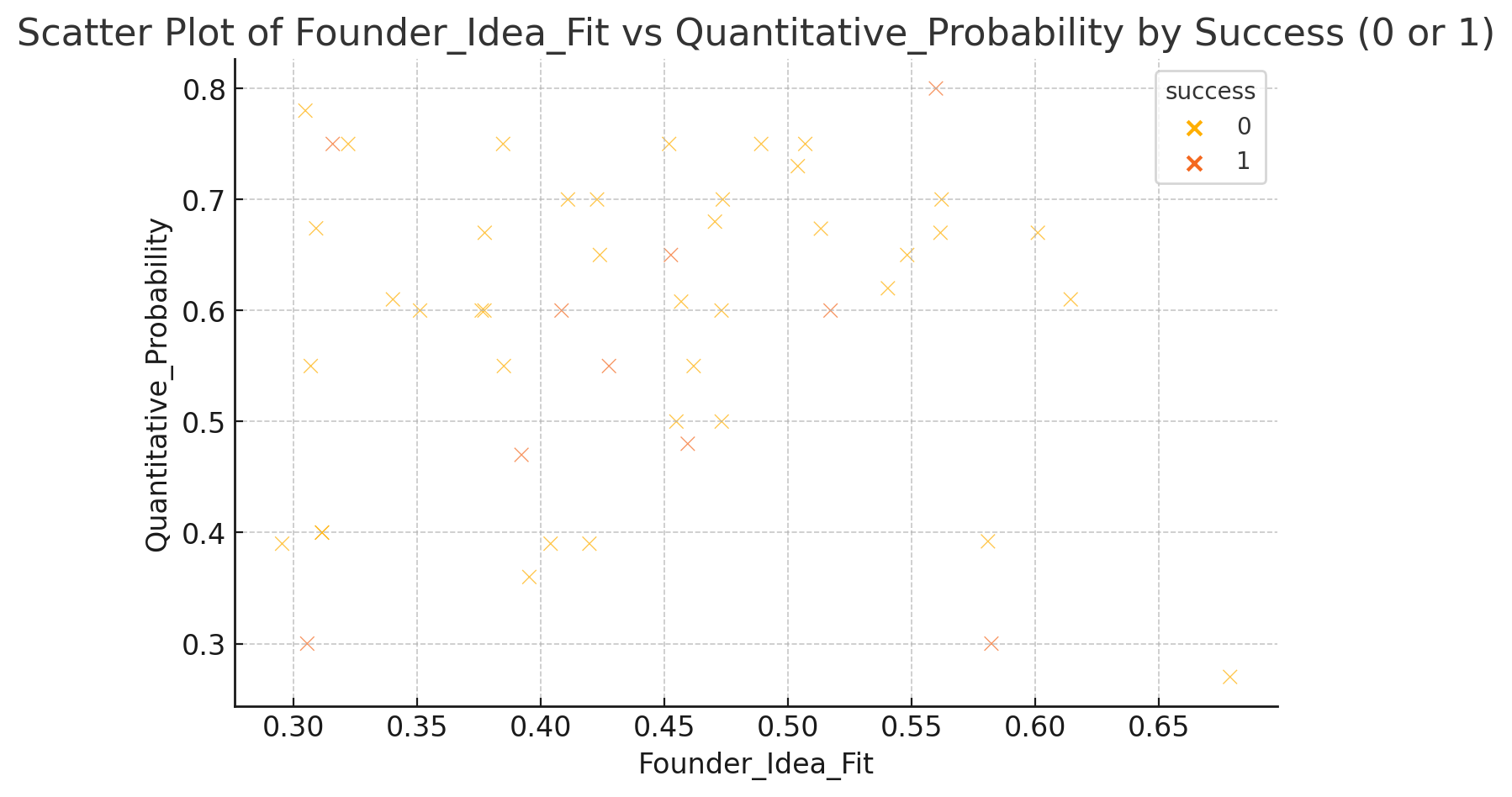}
    \caption{Scatter Plot of Founder Idea Fit vs Quantitative Probability by Success (0 or 1)}
    \label{fig:enter-label}
\end{figure}

\begin{figure}
    \centering
    \includegraphics[width=1\linewidth]{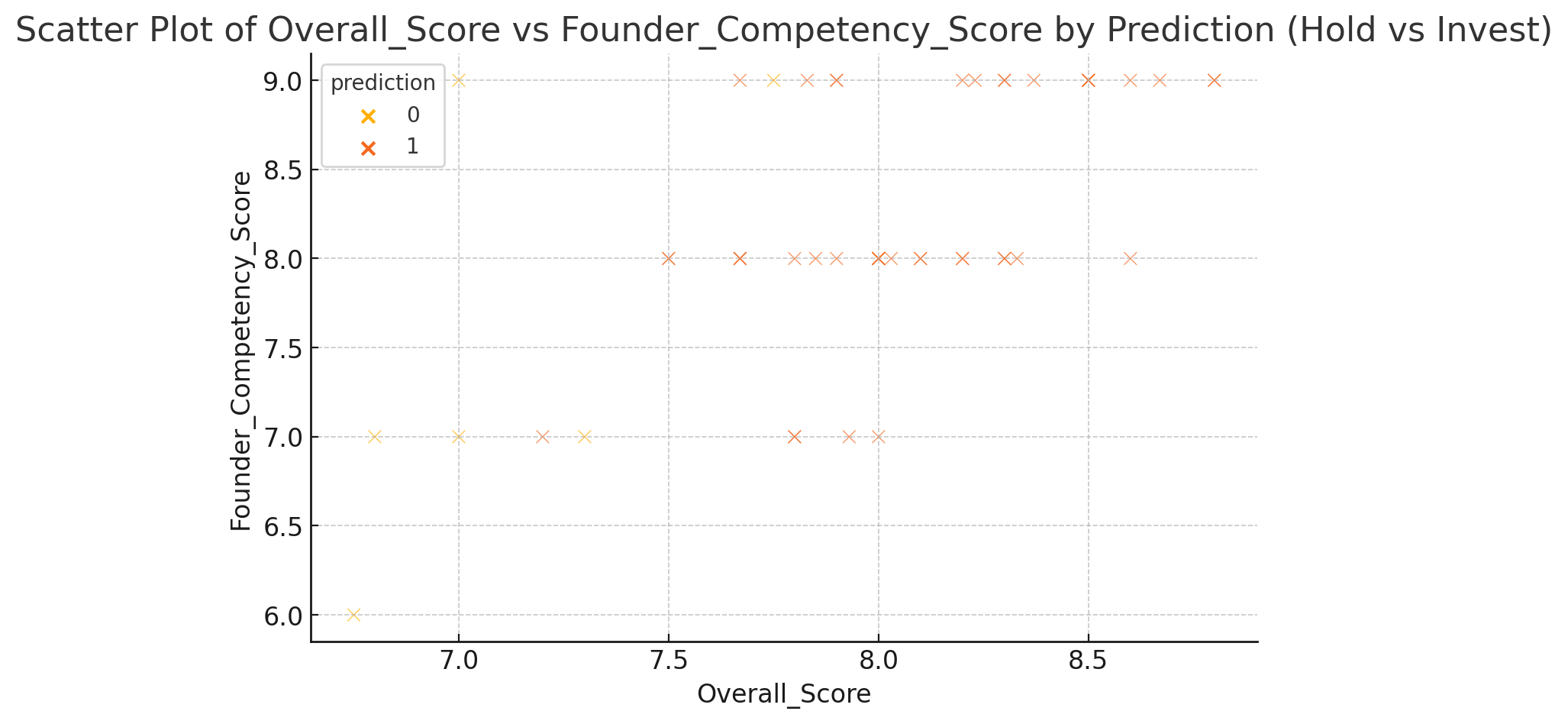}
    \caption{Scatter Plot of Founder Idea Fit vs Quantitative Probability by Prediction (Hold vs Invest}
    \label{fig:enter-label}
\end{figure}

\begin{figure}
    \centering
    \includegraphics[width=1\linewidth]{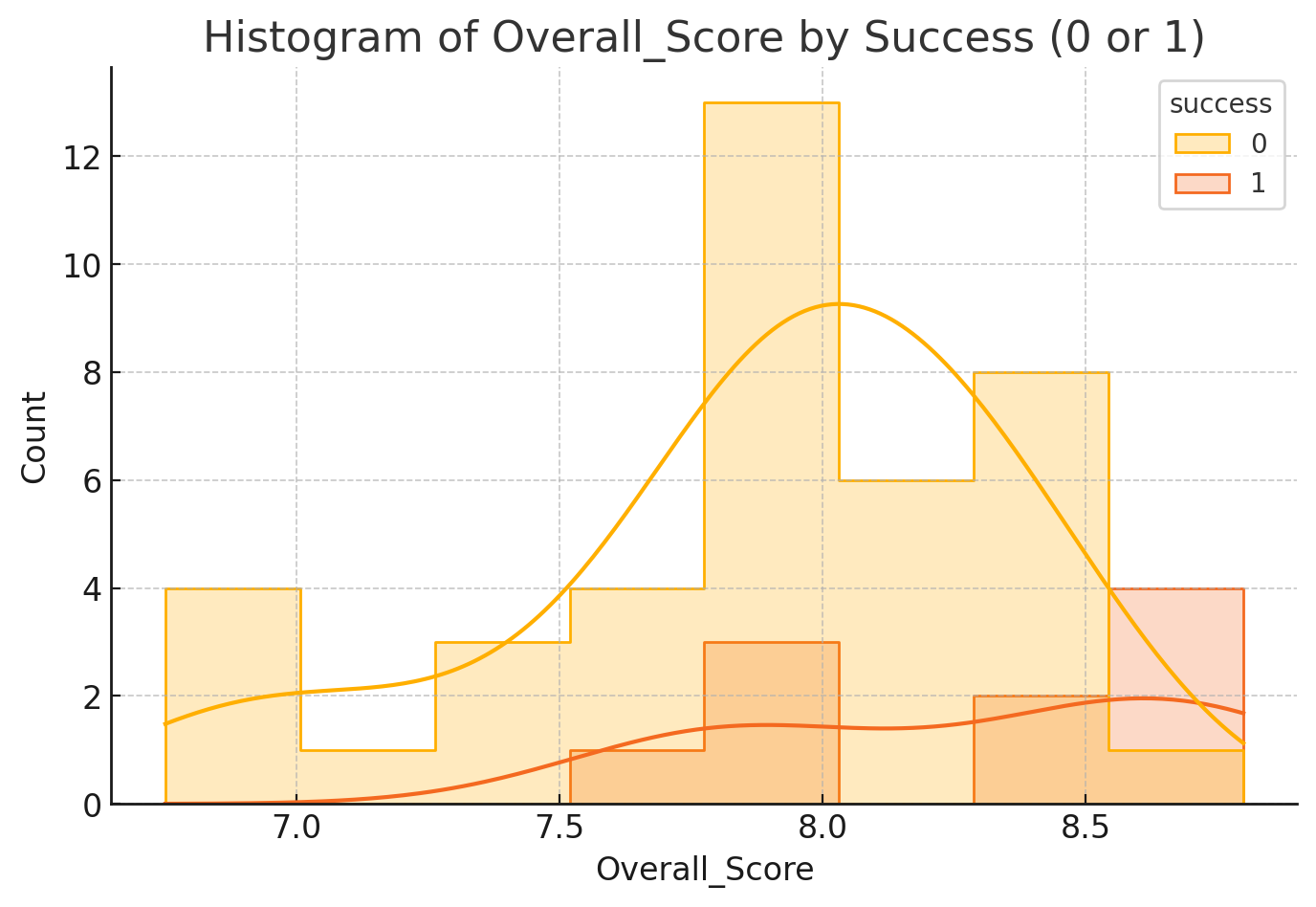}
    \caption{Histogram of Overall Score by Success (0 or 1)}
    \label{fig:enter-label}
\end{figure}

\begin{figure}
    \centering
    \includegraphics[width=1\linewidth]{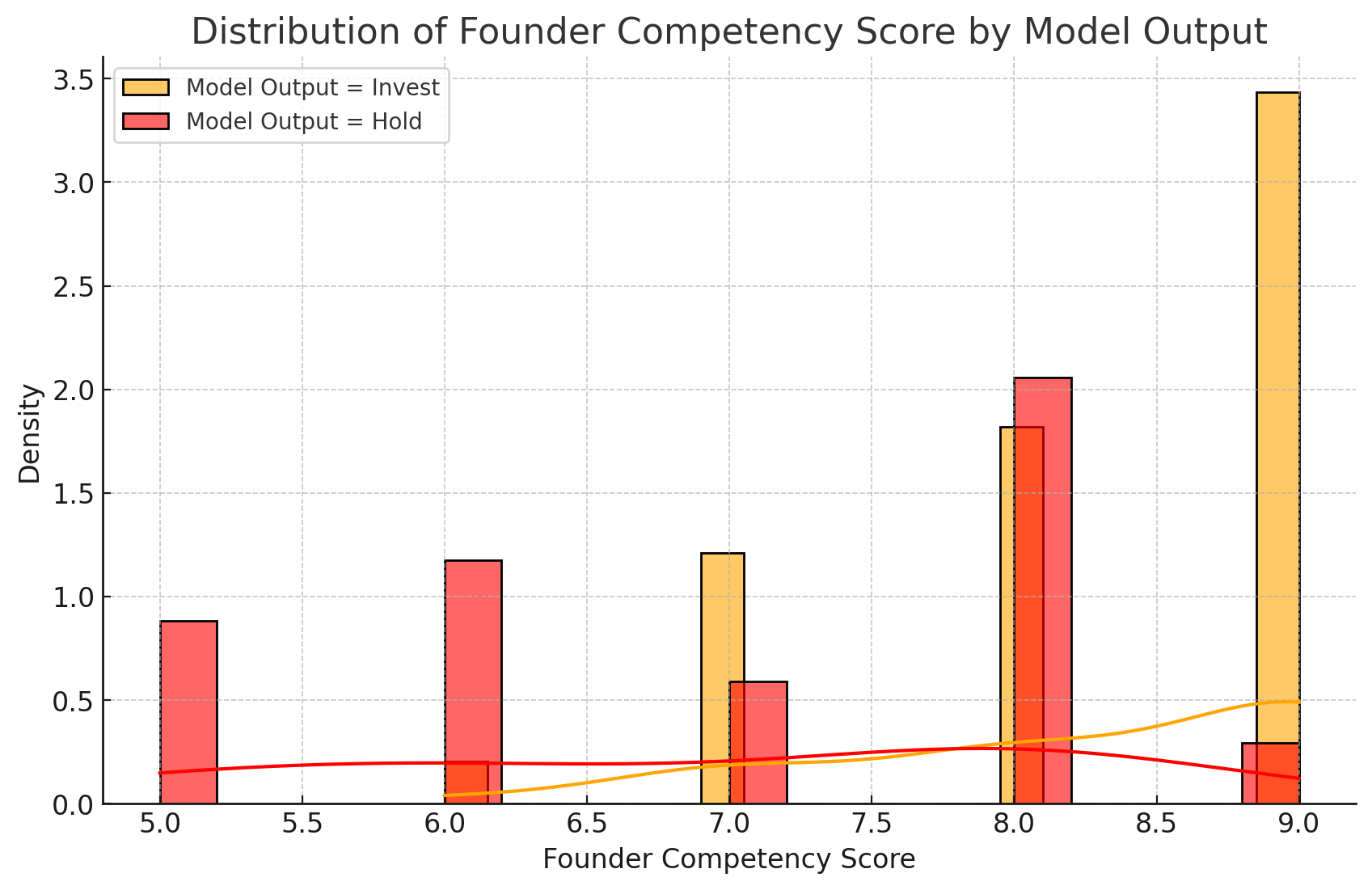}
    \caption{Distribution of Founder Competency Score by Model Output}
    \label{fig:enter-label}
\end{figure}

\subsection {Sample Numerical Output}

A table is listed here as a sample summary. This is independent showcase of summary.

\begin{table}[ht]
\centering
\begin{tabular}{@{}lp{3cm}@{}}
\toprule
\textbf{Metric} & \textbf{Value} \\ 
\midrule
Overall Prediction & Successful, 85\% (overall positive) \\ 
Founder Segmentation & L5 (indicating high likelihood of success) \\ 
Founder-Idea Fit & 0.58861464 (indicating a good fit) \\ 
Categorical Prediction & Successful \\ 
Market Viability Score & 8 (strong market viability) \\ 
Product Viability Score & 8 (strong product viability) \\ 
Founder Competency Score & 9.25 (strong founder competency) \\ 
\bottomrule
\end{tabular}
\caption{Sample Numerical SSFF Outputs \& Implications}
\label{tab:evaluation_metrics}
\end{table}

\subsection {Case Studies Summary}

The below case is for a specific series-A company and its founders. The same input is given to both SSFF and Vanilla GPT. 

\subsubsection {Version 1: SSFF Analysis Summary}

\textbf{Founder's Competency \& Rating}
\begin{itemize}
    \item redacted: Co-Founder \& CEO at Company (redacted), MBA from Cambridge, featured in business magazines, active user engagement in 50 countries. Rating: L3 - Moderate likelihood of success.
    \item redacted: CTO at Company (redacted), Ph.D. from UC Berkeley, extensive experience in tech. Rating: L3 - Moderate likelihood of success.
    \item Combined Rating: L3, indicating moderate likelihood of success for both founders.
\end{itemize}

\textbf{Product Viability \& Competency}
\begin{itemize}
    \item Viability Score: 9/10.
    \item High user satisfaction and active engagement in 50 countries.
    \item Market Fit: Addresses sleep deprivation among parents, uses AI and expert insights for personalized solutions.
    \item Scalability: Potential expansion into broader child wellness sectors.
\end{itemize}

\textbf{Market Viability}
\begin{itemize}
    \item Viability Score: 8.5/10.
    \item Growth Rates: Positive trends in children’s wellness markets.
    \item Market Timing: Growing demand for digital sleep solutions.
    \item Strategy: Strong digital presence, content marketing, subscription models.
\end{itemize}

\textbf{News \& Market Background}
\begin{itemize}
    \item Recent funding of \$12.5 million.
    \item High public sentiment and user satisfaction.
    \item Market Sentiment: Strong, driven by technological advancements in child wellness.
\end{itemize}

\textbf{Quantifiable Modelling}
\begin{itemize}
    \item Founder-Idea Fit Score: 0.41 (Moderate).
    \item Predicted Outcome: 35\% success probability.
    \item Recommendation: Hold on investments; monitor market and founders' performance.
\end{itemize}

\textbf{Recommendation:} \textit{Hold}
\begin{itemize}
    \item \textbf{Strengths:} Strong market and product viability, innovative approach, positive user feedback, and substantial funding.
    \item \textbf{Concerns:} Moderate founder competency and fit with the startup idea, potential gaps in child wellness expertise.
    \item \textbf{Action:} Monitor the startup’s progress, particularly in addressing identified gaps and demonstrating traction in the market. Reconsider investment after close monitoring and ensuring the founders’ readiness to bridge the gaps.
\end{itemize}

\begin{figure}
    \centering
    \includegraphics[width=1\linewidth]{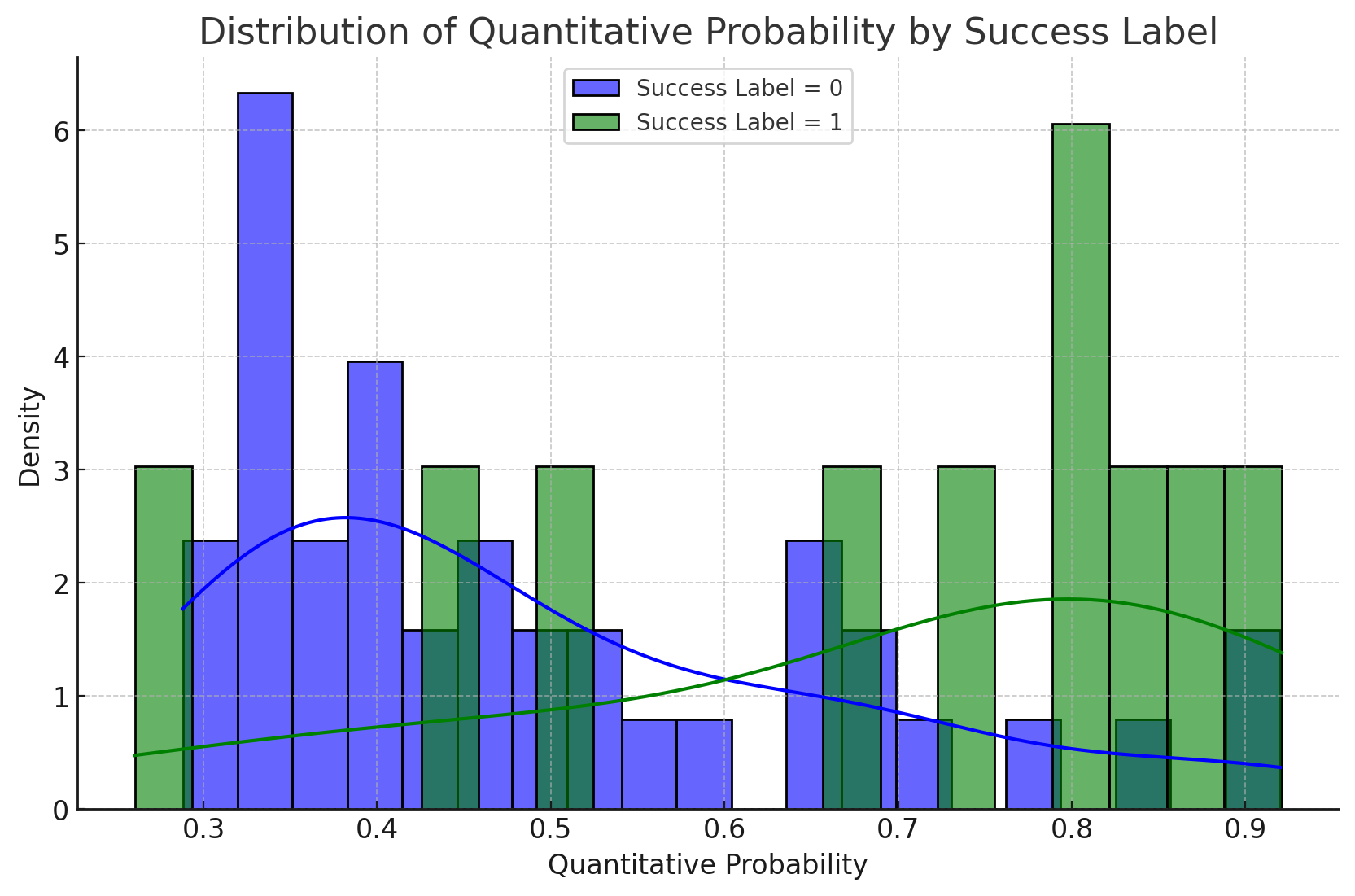}
    \caption{Distribution of Quantitative Probability (Level of Confidence) by Success Label}
    \label{fig:enter-label}
\end{figure}

\subsubsection {Version 2: Vanilla GPT-4o Version Analysis Summary}

\textbf{Founder's Competency}

\textbf{redacted (Co-Founder \& CEO)}
\begin{itemize}
    \item \textbf{Experience:} Over seven years leading Company (redacted), with prior roles in project management and client solutions.
    \item \textbf{Education:} MBA from the University of Cambridge/ 
\end{itemize}

\textbf{redacted (Co-Founder \& CTO)}
\begin{itemize}
    \item \textbf{Experience:} Co-founder of Company (redacted) and Company (redacted), with a background in low-power wireless IP implementation.
    \item \textbf{Education:} Ph.D. in Electrical Engineering and Computer Sciences from UC Berkeley. 
\end{itemize}

\textbf{Rating:} Strong technical and managerial expertise, with significant experience in relevant fields.

\textbf{Product Viability}
\begin{itemize}
    \item Strong product-market fit, leveraging AI and expert consultation for unique value.
    \item High user satisfaction, indicating acceptance and efficacy.
\end{itemize}

\textbf{Market Viability}
\begin{itemize}
    \item Large potential in child sleep solutions.
    \item Well-positioned in the digital health sector, and aligned with the current trends.
\end{itemize}

\textbf{News \& Market Background}
\begin{itemize}
    \item Growing interest in digital health solutions.
    \item Need to navigate regulatory landscapes and competitive pressures.
\end{itemize}

\textbf{Quantifiable Modeling}
\begin{itemize}
    \item Strong foundational backgrounds of founders.
    \item Significant market opportunities in child wellness.
    \item Recommendations: Invest in R\&D, market expansion, and strategic partnerships.
\end{itemize}

\textbf{Overall Investment Recommendation}
Company (redacted) presents a compelling investment opportunity. Strategic investments in AI, market expansion, and partnerships are advised to capitalize on its innovative approach and strong market presence.

\subsection{Declaration of AI Assistance Usage}

AI assistants are helpful in our workflow, streamlining code structuring, facilitating efficient paper reading, and providing robust grammatical revisions to ensure clarity and precision in our work.

\end{document}